\documentclass[runningheads]{llncs}
\usepackage{graphicx}
\usepackage{amsmath,amssymb} 
\usepackage{color}
\usepackage{subfigure}
\usepackage{bm}
\usepackage[width=122mm,left=12mm,paperwidth=146mm,height=193mm,top=12mm,paperheight=217mm]{geometry}
\usepackage{algorithm}
\usepackage{algorithmic}
\usepackage{booktabs}
\newcommand*{\affaddr}[1]{#1} 
\newcommand*{\affmark}[1][*]{\textsuperscript{#1}}

\begin{document}
\newcommand{\doublequote}[1]{``#1''}
\newcommand{\bsym}[1]{\boldsymbol #1}


\pagestyle{headings}
\mainmatter
\def\ECCV18SubNumber{}  
\title{Open Set Domain Adaptation by Backpropagation}
\authorrunning{Saito, Yamamoto, Ushiku and Harada}
\titlerunning{Open Set Domain Adaptation by Backpropagation} 
\author{%
  Kuniaki Saito\affmark[1], Shohei Yamamoto\affmark[1], Yoshitaka Ushiku\affmark[1], and Tatsuya Harada\affmark[1,2]\\
  \affaddr{\affmark[1]The University of Tokyo}, \affaddr{\affmark[2]RIKEN}\\
  \tt\small\{k-saito, yamamoto, ushiku, harada\}@mi.t.u-tokyo.ac.jp}


\institute{}

\maketitle

\begin{abstract}
Numerous algorithms have been proposed for transferring knowledge from a label-rich domain (source) to a label-scarce domain (target).
Almost all of them are proposed for closed-set scenario, where the source and the target domain completely share the class of their samples. We call the shared class the \doublequote{known class.}
However, in practice, when samples in target domain are not labeled, we cannot know whether the domains share the class. A target domain can contain samples of classes that are not shared by the source domain.
We call such classes the \doublequote{unknown class} and algorithms that work well in the open set situation are very practical. However, most existing distribution matching methods for domain adaptation do not work well in this setting because unknown target samples should not be aligned with the source.

In this paper, we propose a method for an open set domain adaptation scenario which utilizes adversarial training. 
A classifier is trained to make a boundary between the source and the target samples whereas a generator is trained to make target samples far from the boundary. Thus, we assign two options to the feature generator: aligning them with source known samples or rejecting them as unknown target samples. 
This approach allows to extract features that separate unknown target samples from known target samples. 
Our method was extensively evaluated on domain adaptation setting and outperformed other methods with a large margin in most setting.

\end{abstract}

\section{Introduction}
Deep neural networks have demonstrated significant performance on many image recognition tasks~\cite{krizhevsky2012imagenet}. One of the main problems of such methods is that basically, they cannot recognize samples as unknown, whose class is absent during training. We call such a class as an \doublequote{unknown class} and the categories provided during training is referred to as the \doublequote{known class.} If these samples can be recognized as unknown, we can arrange noisy datasets and pick out the samples of interest from them. Moreover, if robots working in the real-world can detect unknown objects and ask annotators to give labels to them, these robots will be able to easily expand their knowledge. Therefore, the open set recognition is a very important problem.

\begin{figure}[t]
\centering
\includegraphics[width=\hsize]{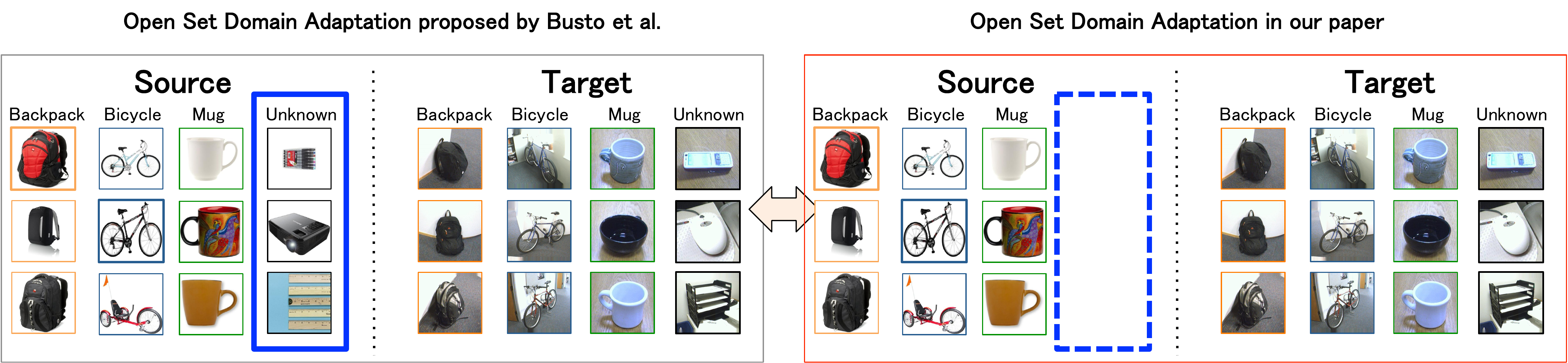}
\caption{A comparison between existing open set domain adaptation setting and our setting. {\bf Left}: Existing setting of open set domain adaptation~\cite{busto2017open}. It is assumed that access is granted to the unknown source samples although the class of unknown source does not overlap with that of unknown target. {\bf Right}: Our setting. We do not assume the accessibility to the unknown samples in the source domain. We propose a method that can be applied even when such samples are absent.}
\label{fig:intro}
\end{figure}

In domain adaptation, we aim to train a classifier from a label-rich domain (source domain) and apply it to a label-scarce domain (target domain). Samples in different domains have diverse characteristics which degrade the performance of a classifier trained in a different domain. 
Most works on domain adaptation assume that samples in the target domain necessarily belong to the class of the source domain. However, this assumption is not realistic. Consider the setting of an unsupervised domain adaptation, where only unlabeled target samples are provided. We cannot know that the target samples necessarily belong to the class of the source domain because they are not given labels. Therefore, open set recognition algorithm is also required in domain adaptation.
For this problem, the task called open set domain adaptation was recently proposed~\cite{busto2017open} where the target domain contains samples that do not belong to the class in the source domain as shown in the left of Fig. \ref{fig:intro}. 
The solution to the problem should allow to classify unknown target samples as "unknown" and to classify known target samples into correct known categories.
They~\cite{busto2017open} utilized unknown source samples to classify unknown target samples as unknown. However, collecting unknown source samples is also expensive because we must collect diverse and many unknown source samples to obtain the concept of \doublequote{unknown.} 
Then, in this paper, we present a more challenging open set domain adaptation (OSDA) that does not provide any unknown source samples, and we propose a method for it. That is, we propose a method where we have access to only known source samples and unlabeled target samples for open set domain adaptation as shown in the right of Fig. \ref{fig:intro}. 

How can we solve the problem? We think that there are mainly two problems. First, in this situation, we do not have knowledge about which samples are the unknown samples. Thus, it seems difficult to delineate a boundary between known and unknown classes. The second problem is related to the domain's difference. Although we need to align target samples with source samples to reduce this domain's difference, unknown target samples cannot be aligned due to the absence of unknown samples in the source domain. The existing distribution matching method is aimed at matching the distribution of the target with that of the source. However, this method cannot be applied to our problem. In OSDA, we must reject unknown target samples without aligning them with the source.


To solve the problems, we propose a new approach of adversarial learning that enables generator to separate target samples into known and unknown classes. A comparison with existing methods is shown in Fig. \ref{fig:comparison_method}. Unlike the existing distribution alignment methods that only match the source and target distribution, our method facilitates the rejection of unknown target samples with high accuracy as well as the alignment of known target samples with known source samples. We assume that we have two players in our method, i.e., the feature generator and the classifier. The feature generator generates features from inputs, and the classifier takes the features and outputs $K+1$ dimension probability, where $K$ indicates the number of known classes. The $K+1$ th dimension of output indicates the probability for the unknown class. The classifier is trained to make a boundary between source and target samples whereas the feature generator is trained to make target samples far from the boundary. 
Specifically, we train the classifier to output probability $t$ for unknown class, where $0 < t < 1$. We can build a decision boundary for unknown samples by weakly training a classifier to classify target samples as unknown. 
To deceive the classifier, the feature generator has two options to increase or to decrease the probability. As such, we assign two options to the feature generator: aligning them with samples in the source domain or rejecting them as unknown. 

The contribution of our paper is as follows.
\vspace{-3mm}
\begin{enumerate}
\item We present the open set domain adaptation where unknown source samples are not provided. The setting is more challenging than the existing setting.
\item We propose a new adversarial learning method for the problem. The method enables training of the feature generator to learn representations which can separate unknown target samples from known ones.
\item We evaluate our method on adaptation for digits and objects datasets and demonstrate its effectiveness. Additionally, the effectiveness of our method was demonstrated in standard open set recognition experiments where we are provided unlabeled unknown samples during training. 
\end{enumerate}

\begin{figure}[t]
\subfigure[]{
   \includegraphics[width=35mm]{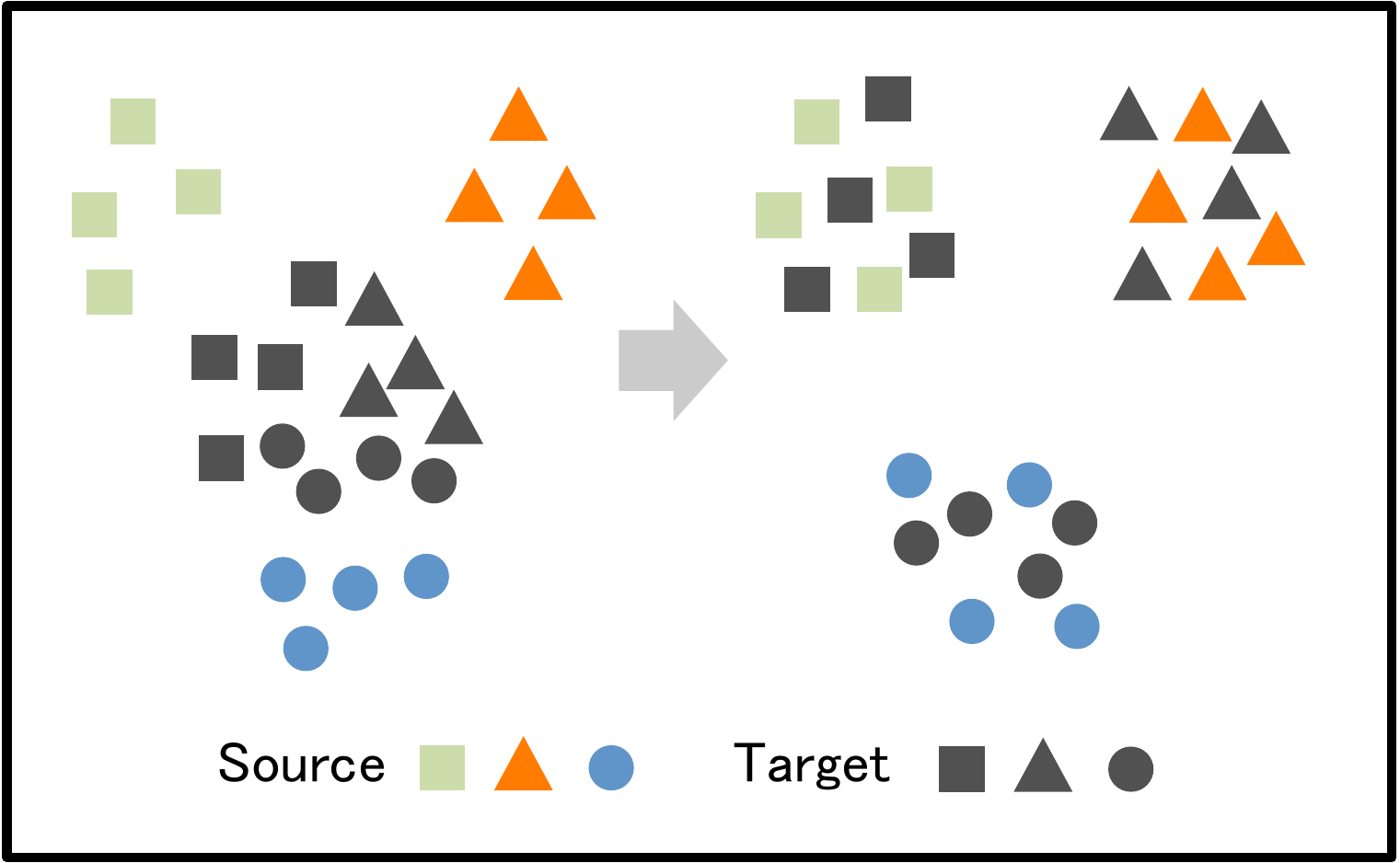}\label{fig:closed_da}}
 \subfigure[]{
  \includegraphics[width=38mm]{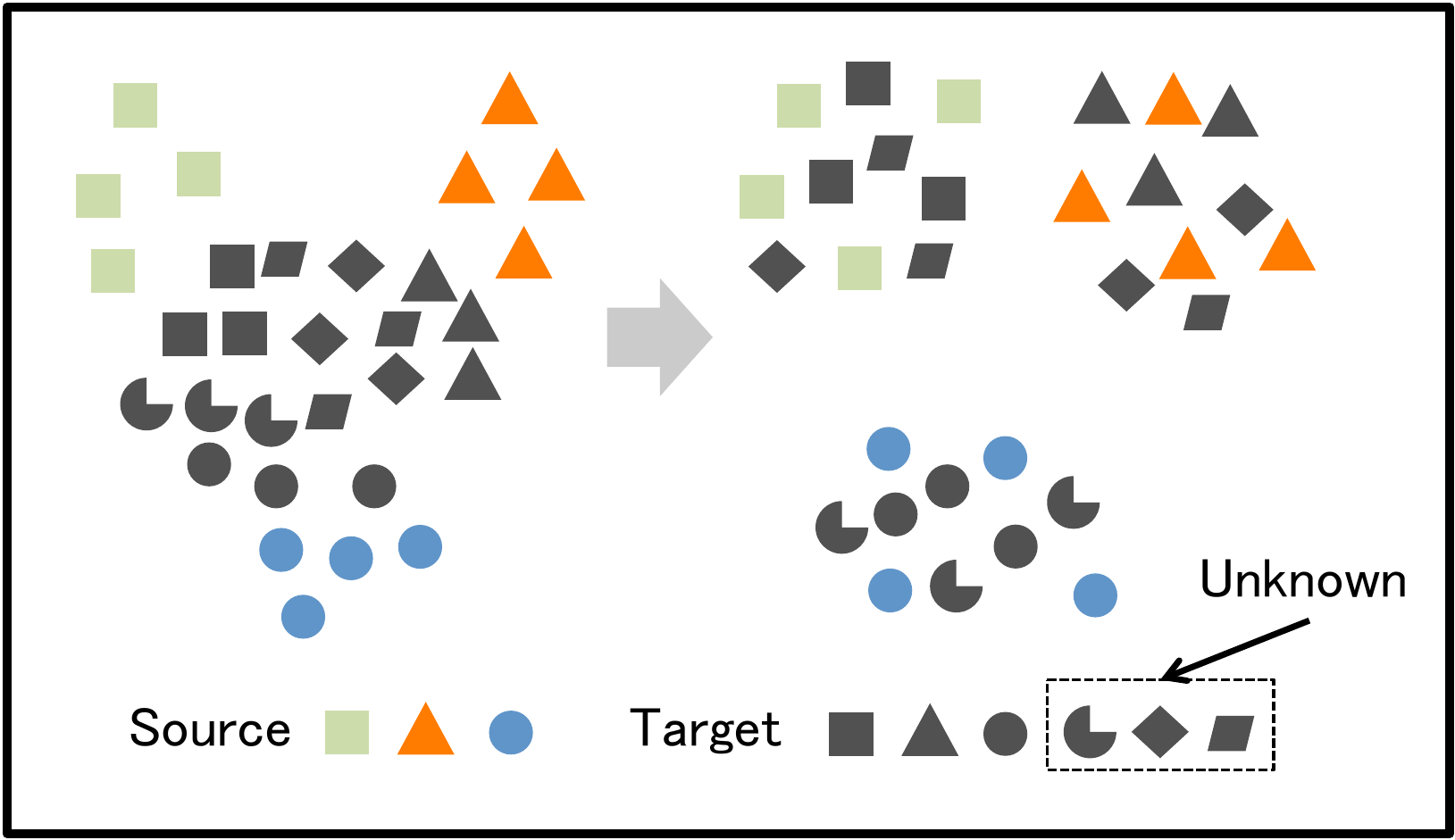}\label{fig:open_da}}
 \subfigure[]{
  \includegraphics[width=38mm]{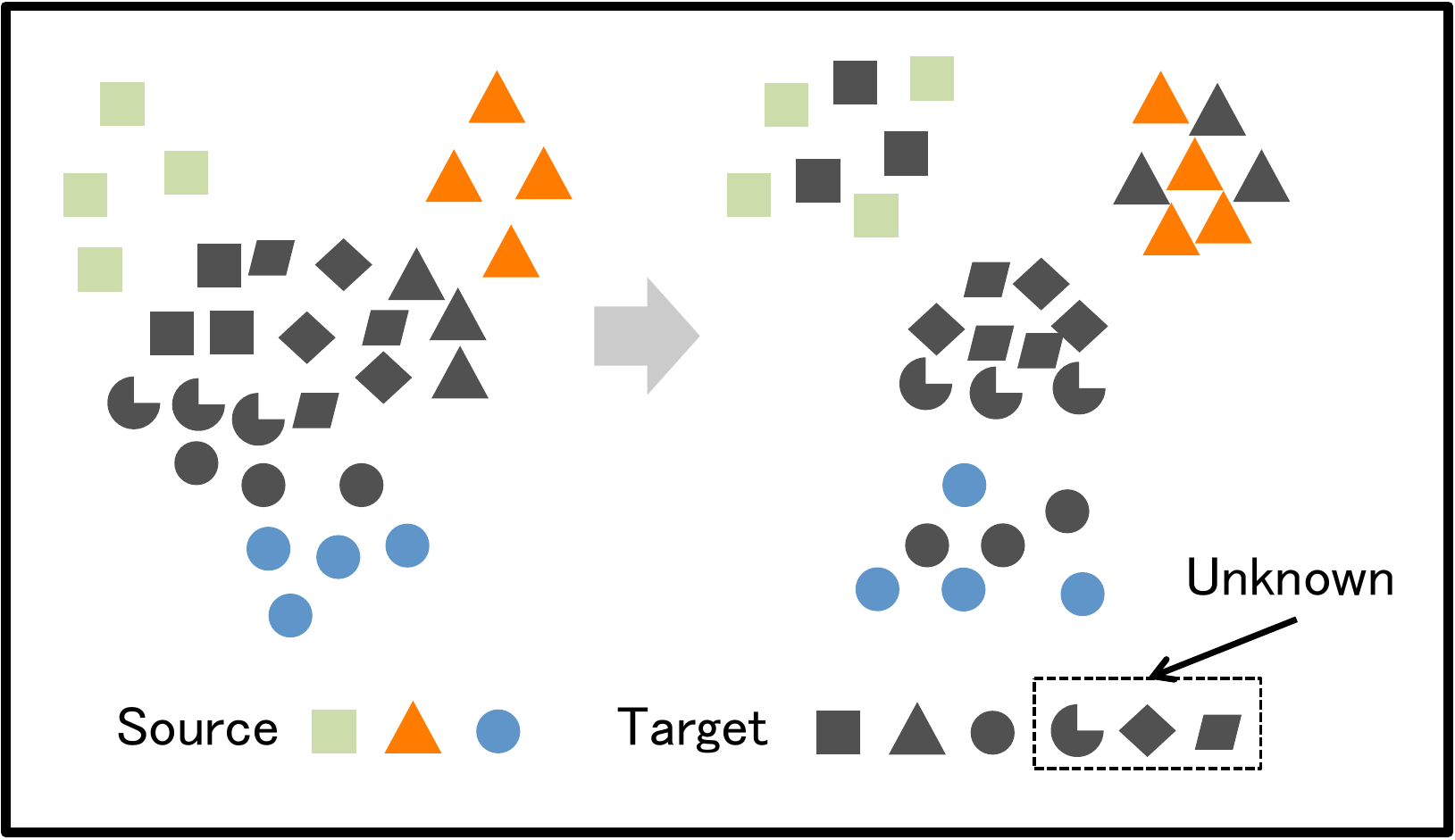}\label{fig:open_ours}}
  \label{fig:comparison_method}
  \vspace{-3mm}
  \caption{\subref{fig:closed_da}: Closed set domain adaptation with distribution matching method. \subref{fig:open_da}: Open set domain adaptation with distribution matching method. Unknown samples are aligned with known source samples. \subref{fig:open_ours}: Open set domain adaptation with our proposed method. Our method enables to learn features that can reject unknown target samples.}
  \label{fig:comparison_method}
\end{figure}
\vspace{-5mm}
\section{Related Work}
\vspace{-3mm}
In this section, we briefly introduce methods for domain adaptation and open set recognition.
\vspace{-5mm}
\subsection{Domain Adaptation}
\vspace{-3mm}
Domain adaptation for image recognition has attracted attention for transferring the knowledge between different domains and reducing the cost for annotating a large number of images in diverse domains. Benchmark datasets are released~\cite{saenko2010adapting}, and many methods for unsupervised domain adaptation and semi-supervised domain adaptation have been proposed~\cite{ganin2014unsupervised,gong2013connecting,long2015learning,long2016unsupervised,gong2012geodesic,saito2017asymmetric,sener,ghifary2016deep}.
As previously indicated, unsupervised and semi-supervised domain adaptation focus on the situation where different domains completely share the class of their samples, which may not be practical especially in unsupervised domain adaptation.

One of the effective methods for unsupervised domain adaptation are distribution matching based methods~\cite{ganin2014unsupervised,long2015learning,tzeng2014deep,bousmalis2016unsupervised,saito2017maximum}.
Each domain has unique characteristics of their features, which decrease the performance of classifiers trained on a different domain. Therefore, by matching the distributions of features between different domains, they aim to extract domain-invariantly discriminative features. This technique is widely used in training neural networks for domain adaptation tasks~\cite{ganin2014unsupervised,hoffman2016fcns}. The representative of the methods harnesses techniques used in Generative Adversarial Networks (GAN)~\cite{goodfellow2014generative}. GAN trains a classifier to judge whether input images are fake or real images whereas the image generator is trained to deceive it. In domain adaptation, similar to GAN, the classifier is trained to judge whether the features of the middle layers are from a target or a source domain whereas the feature generator is trained to deceive it. Variants of the method and extensions to the generative models for domain adaptation have been proposed~\cite{bousmalis2016unsupervised,bousmalis2016domain,tzeng2017adversarial,taigman2016unsupervised,liu2017unsupervised}. 
Maximum Mean Discrepancy (MMD) ~\cite{gretton2007kernel} is also a representative way to measure the distance between domains. The distance is utilized to train domain-invariantly effective neural networks, and its variants are proposed~\cite{long2015learning,long2016unsupervised,long2016deep,yan2017mind}.

The problem is that these methods do not assume that the target domain has categories that are not included in the source domain. The methods are not supposed to perform well on our open set domain adaptation scenario. This is because all target samples including unknown classes will be aligned with source samples. Therefore, this makes it difficult to detect unknown target samples.

In contrast, our method enables to categorization of unknown target samples as unknown, although we are not provided any labeled target unknown samples during training. We will compare our method with MMD and domain classifier based methods in experiments. We utilize the technique of distribution matching methods technique to achieve open set recognition. However, the main difference is that our method allows the feature generator to reject some target samples as outliers. 
\vspace{-3mm}
\subsection{Open Set Recognition}
A wide variety of research has been conducted to reject outliers while correctly classifying inliers during testing. 

Multi-class open set SVM is proposed by~\cite{jain2014multi}. They propose to reject unknown samples by learning SVMs that assign probabilistic decision scores. The aim is to reject unknown samples using a threshold probability value. In addition, method of harnessing deep neural networks for open set recognition was proposed~\cite{bendale2016towards}. They introduced OpenMax layer, which estimates the probability of an input being from an unknown
class. Moreover, to give supervision of the unknown samples, a method to generate these samples was proposed~\cite{ge2017generative}. The method utilizes GAN to generate unknown samples and use it to train neural networks, then combined it with OpenMax layer. 
In order to recognize unknown samples as unknown during testing, these methods defined a threshold value to reject unknown samples. Also, they do not assume that they can utilize unlabeled samples including known and unknown classes during training. 

In our work, we propose a method that enables us to deal with the open set recognition problem in the setting of the domain adaptation. In this setting, the distribution of the known samples in the target domain is different from that of the samples in the source domain, which makes the task more difficult.

\begin{figure}[t]
\centering
\includegraphics[height=0.32\hsize]{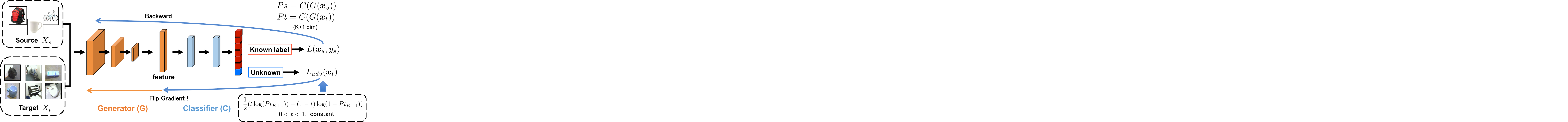}
\vspace{-5mm}
\caption{The proposed method for open set domain adaptation. Classifier networks output $K+1$ dimensional probabilistic output. The network is trained to correctly classify source samples. For target samples, the classifier is trained to output $t$ for the probability of the unknown class whereas the generator is trained to deceive it. We utilize the gradient reversal layer for the adversarial training.}
\label{fig:pipeline}
\end{figure}
\vspace{-3mm}
\section{Method}
\vspace{-3mm}
First, we provide an overall overview of our method, then we explain the actual training procedure and provide an analysis of our method by comparing it with existing open set recognition algorithm. The overview of our method is shown in Fig. \ref{fig:pipeline}.
\vspace{-3mm}
\subsection{Problem Setting and Overall Idea}
\vspace{-2mm}
We assume that a labeled source image $\bsym{x_{s}}$ and a corresponding label $y_{s}$ drawn from a set of labeled source images \{$X_{s},Y_{s}$\} are available,
as well as an unlabeled target image $\bsym{x_{t}}$ drawn from unlabeled target images $X_{t}$. The source images are drawn only from known classes whereas target images can be drawn from unknown class. 
In our method, we train a feature generation network $G$, which takes inputs $\bsym{x_{s}}$ or $\bsym{x_{t}}$, and a network $C$, which takes features from $G$ and classifies them into $K+1$ classes, where the $K$ denotes the number of known categories. Therefore, $C$ outputs a $K+1$-dimensional vector of logits \{$l_{1},l_{2},l_{3}...l_{K+1}$\} per one sample. 

The logits are then converted to class probabilities by applying the softmax function. Namely, the probability that $\bsym{x}$ is classified into class $j$ is denoted by $p(y=j|\bsym{x})=\frac{\exp(l_{j})}{\sum_{k=1}^{K+1}\exp(l_{k})}$. 1 $\sim$ K dimensions indicate the probability for the known classes whereas $K+1$ dimension indicates that for the unknown class.
We use the notation $p(\bsym{y}|\bsym{x})$ to denote the $K+1$-dimensional probabilistic output for input $\bsym{x}$. 

Our goal is to correctly categorize known target samples into  corresponding known class and recognize unknown target samples as unknown. 
We have to construct a decision boundary for the unknown class, although we are not given any information about the class. Therefore, we propose to make a pseudo decision boundary for unknown class by weakly training a classifier to recognize target samples as unknown class. Then, we train a feature generator to deceive the classifier. 
The important thing is that feature generator has to separate unknown target samples from known target samples. If we train a classifier to output $p(y=K+1|{\bsym{x_t}}) = 1.0$ and train the generator to deceive it, then ultimate objective of the generator is to completely match the distribution of the target with that of the source. Therefore, the generator will only try to decrease the value of the probability for unknown class.
This method is used for training Generative Adversarial Networks for semi-supervised learning~\cite{salimans2016improved} and should be useful for unsupervised domain adaptation. However, this method cannot be directly applied to separate unknown samples from known samples. 

Then, to solve the difficulty, we propose to train the classifier to output $p(y=K+1|{\bsym{x_t}}) = t$, where $0<t<1$. We train the generator to deceive the classifier. That is, the objective of the generator is to maximize the error of the classifier. In order to increase the error, the generator can choose to increase the value of the probability for an unknown class, which means that the sample is rejected as unknown. 
For example, consider when $t$ is set as a very small value, it should be easier for generator to increase the probability for an unknown class than to decrease it to maximize the error of the classifier. Similarly, it can choose to decrease it to make $p(y=K+1|{\bsym{x_t}})$ lower than $t$, which means that the sample is aligned with source. 
In summary, the generator will be able to choose whether a target sample should be aligned with the source or should be rejected. In all our experiments, we set the value of $t$ as 0.5. If $t$ is larger than 0.5, the sample is necessarily recognized as unknown. Thus, we assume that this value can be a good boundary between known and unknown. In our experiment, we will analyze the behavior of our model when this value is varied. 
\vspace{-3mm}
\subsection{Training Procedure}
\vspace{-3mm}
We begin by demonstrating how we trained the model with our method.
First, we trained both the classifier and the generator to categorize source samples correctly. We use a standard cross-entropy loss for this purpose. 
\newcommand{\mymin}{\mathop{\rm min}\limits}
\newcommand{\mymax}{\mathop{\rm max}\limits}
\newcommand{\1}{\mbox{1}\hspace{-0.25em}\mbox{l}}
\begin{eqnarray}
L_s({\bsym{x_s}},y_s) &=& -\log(p(y=y_s|{\bsym{x_s}}))\\
p(y=y_s|{\bsym{x_s}}) &=& (C \circ G({\bsym{x_s}}))_{y_s}
\end{eqnarray}
In order to train a classifier to make a boundary for an unknown sample, we propose to utilize a binary cross entropy loss. 
\begin{eqnarray}
\label{eq:loss_adv}
L_{adv}(\bsym{x_t}) = -t\log(p(y=K+1|\bsym{x_t})) - (1-t) \log(1-p(y=K+1|\bsym{x_t}))
\end{eqnarray}
, where $t$ is set as 0.5 in our experiment. 
The overall training objective is,
\begin{eqnarray}
\label{eq:update_c}
\mymin_{C} L_s(\bsym{x_s},y_s) + L_{adv}(\bsym{x_t})
\end{eqnarray}
\begin{eqnarray}
\label{eq:update_g}
\mymin_{G} L_s(\bsym{x_s},y_s) - L_{adv}(\bsym{x_t})
\end{eqnarray}
The classifier attempts to set the value of $p(y=K+1|\bsym{x_t})$ equal to $t$ whereas 
the generator attempts to maximize the value of $L_{adv}(\bsym{x_t})$. Thus, it attempts to make the value of $p(y=K+1|\bsym{x_t})$ different from $t$. In order to efficiently calculate the gradient for $L_{adv}(\bsym{x_t})$, we utilize a gradient reversal layer proposed by~\cite{ganin2014unsupervised}. The layer enables flipping of the sign of the gradient during the backward process. Therefore, we can update the parameters of the classifier and generator simultaneously. The algorithm is shown in Alg. \ref{alg:proposed}.
\vspace{-3mm}
\begin{algorithm}[ht]
\caption{\small Minibatch training of the proposed method.}
\begin{algorithmic}

\FOR{the number of training iterations}
  \STATE{$\bullet$ Sample minibatch of $m$ source samples 
  $\bigl\{ \{\bm{x_s},y_s\bigl\}^{(1)}, \dots, \{\bm{x_s},y_s\bigl\}^{(m)} \bigl\}$ from $\{X_s,Y_s\}$.}
    \STATE{$\bullet$ Sample minibatch of $m$ target samples $\{ \bm{x_t}^{(1)}, \dots, \bm{x_t}^{(m)} \}$ from ${X_t}$.}
    \STATE Calculate $L_s(\bsym{x_s},y_s)$ by cross-entropy loss and $L_{adv}(\bsym{x_t})$ following Eq. \ref{eq:loss_adv}.
    \STATE Update the parameter of $G$ and $C$ following Eq. \ref{eq:update_c}, Eq. \ref{eq:update_g}. We used gradient reversal layer for this operation.
  \ENDFOR
\end{algorithmic}
\label{alg:proposed}
\end{algorithm}
\vspace{-5mm}
\subsection{Comparison with Existing Methods}
\vspace{-2mm}
In this section, we analyze our method by comparing it with existing methods for the open set recognition problem. We think that there are three major differences compared to existing methods.
Since most existing methods do not have access to unknown samples during training, they cannot train feature extractors to learn features to reject them. In contrast, in our setting, unknown target samples are included in training samples though we do not know which are unknown samples. Under the condition, our method can train feature extractors to reject unknown samples. 
In addition, existing methods such as open set SVM reject unknown samples if the probability of any known class for a testing sample is not larger than the threshold value. The value is a pre-defined one and does not change across testing samples. However, with regard to our method, we can consider that the threshold value changes across samples because our model assigns different classification outputs to different samples. 
Thirdly, the feature extractor is informed of the pseudo decision boundary between known and unknown classes. Thus, feature extractors can recognize the distance between each target sample and the boundary for the unknown class. It attempts to make it far from the boundary. It makes representations such that the samples similar to the known source samples are aligned with known class whereas ones dissimilar to known source samples are separated from them. 

\vspace{-3mm}
\section{Experiments}
\vspace{-3mm}
We conduct experiments on Office~\cite{saenko2010adapting}, VisDA~\cite{visda} and digits datasets.
\vspace{-3mm}
\subsection{Implementation Detail}
\vspace{-3mm}
We trained the classifier and generator using the features obtained from AlexNet~\cite{krizhevsky2012imagenet} and VGGNet~\cite{simonyan2014very} pretrained on ImageNet~\cite{deng2009imagenet}. In the experiments on both Office and VisDA dataset, we did not update the parameters of the networks. We constructed fully-connected layers with 100 hidden units after the FC8 layers. Batch Normalization~\cite{ioffe2015batch} and Leakly-ReLU layer were employed for stable training. We used momentum SGD with a learning rate $1.0\times10^{-3}$. The momentum was set as $0.9$. The details of the experiments are shown in our supplementary material due to a limit of space.

We implemented three baselines in the experiments. The first baseline is an open set SVM (OSVM)~\cite{jain2014multi}. OSVM utilizes the threshold probability to recognize samples as unknown if the predicted probability is lower than the threshold for any class. We first trained CNN only using source samples, then, use it as a feature extractor. Features are extracted from the output of generator networks when using OSVM. OSVM does not require unknown samples during training. Therefore, we trained OSVM only using source samples and tested them on the target samples.
The second baseline is a combination of Maximum Mean Discrepancy(MMD)~\cite{gretton2007kernel} based training method for neural networks~\cite{long2015learning} and OSVM. MMD is used to match the distribution between different domains in unsupervised domain adaptation. For an open set recognition, we trained the networks with MMD and trained OSVM using the features obtained by the networks. A comparison with this baseline should indicate how our proposed method is different from existing distribution matching methods. The third baseline is a combination of a domain classifier based method, BP~\cite{ganin2014unsupervised} and OSVM. BP is also a representative of a distribution matching method. As was done for MMD, we first trained BP and extracted features to train OSVM. We used the same network architecture to train the baseline models. The experiments were run a total of 3 times for each method, and the average score was reported. We report the standard deviation only in Table \ref{office_20} because of the limit of space. 
\vspace{-3mm}
\subsection{Experiments on Office}

\subsubsection{11 Class Classification}

Firstly, we evaluated our method using Office following the protocol proposed by \cite{busto2017open}. We then compared our method with other existing works. The dataset consists of 31 classes, and 10 classes were selected as shared classes. The classes are also common in the Caltech dataset~\cite{gong2012geodesic}. 
In alphabetical order, 21-31 classes are used as unknown samples in the target domain. The classes 11-20 are used as unknown samples in the source domain in \cite{busto2017open}. However, we did not use it because our method does not require such samples. 
We have to correctly classify samples in the target domain into 10 shared classes or unknown class. In total, 11 class classification was performed in this setting. 
Accuracy averaged over all classes is denoted as OS in all Tables. 
We also show the accuracy measured only on the shared samples of the target domain (OS*(10)). Following \cite{busto2017open}, we show the accuracy averaged over the classes in the OS and OS*. 

We also compared our method with a method proposed by~\cite{busto2017open}. Their method is developed for a situation where unknown samples in the source domain are available. However, they applied their method using OSVM when unknown source samples were absent. In order to better understand the performance of our method, we also show the results which utilized the unknown source samples during training. The values are cited from~\cite{busto2017open}. 

\begin{figure}[h!]
\subfigure[AlexNet]{
   \includegraphics[width=0.17\hsize]{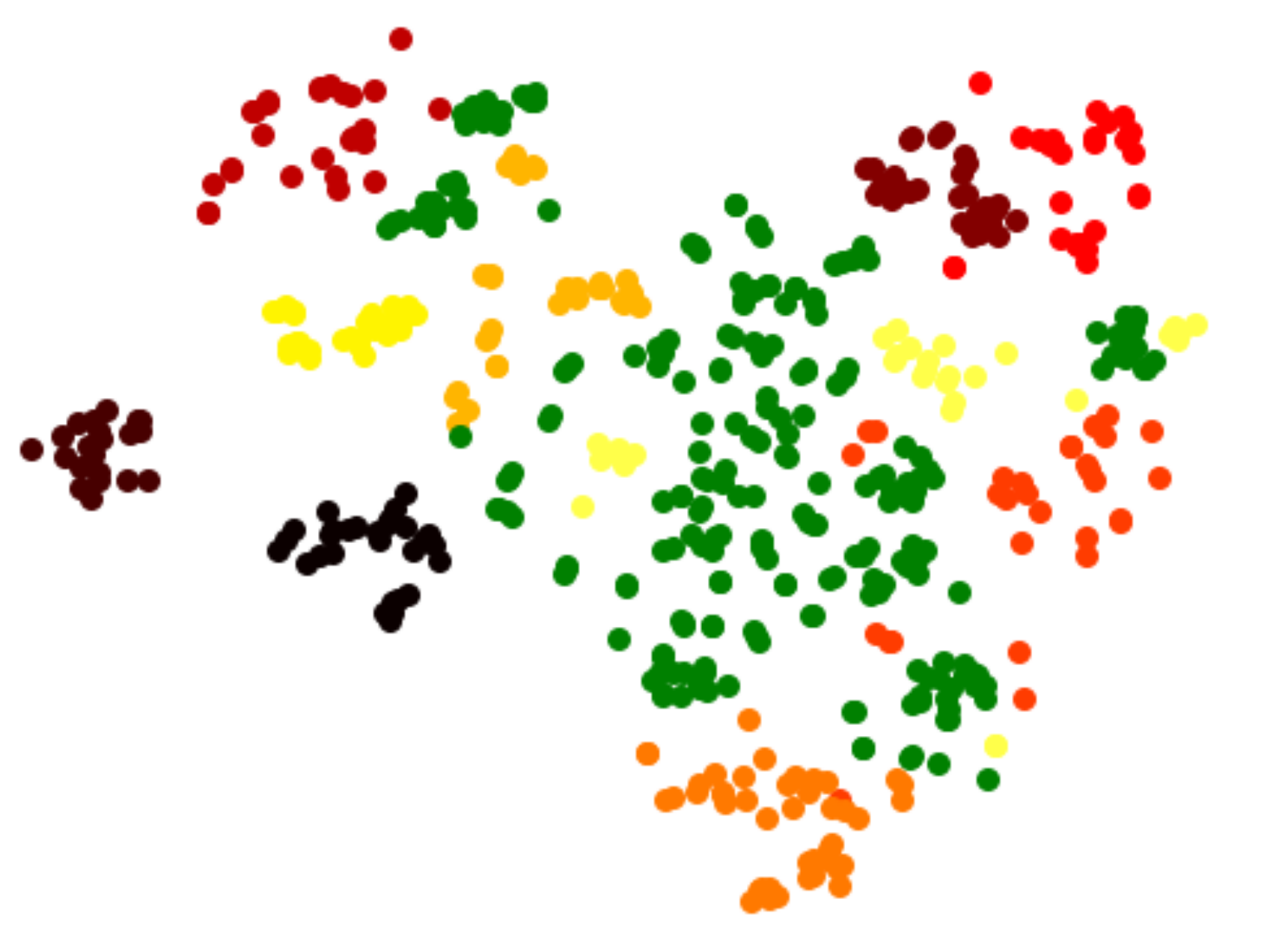}\label{fig:before_adapt}}
\subfigure[\scriptsize{Source Only}]{
   \includegraphics[width=0.17\hsize]{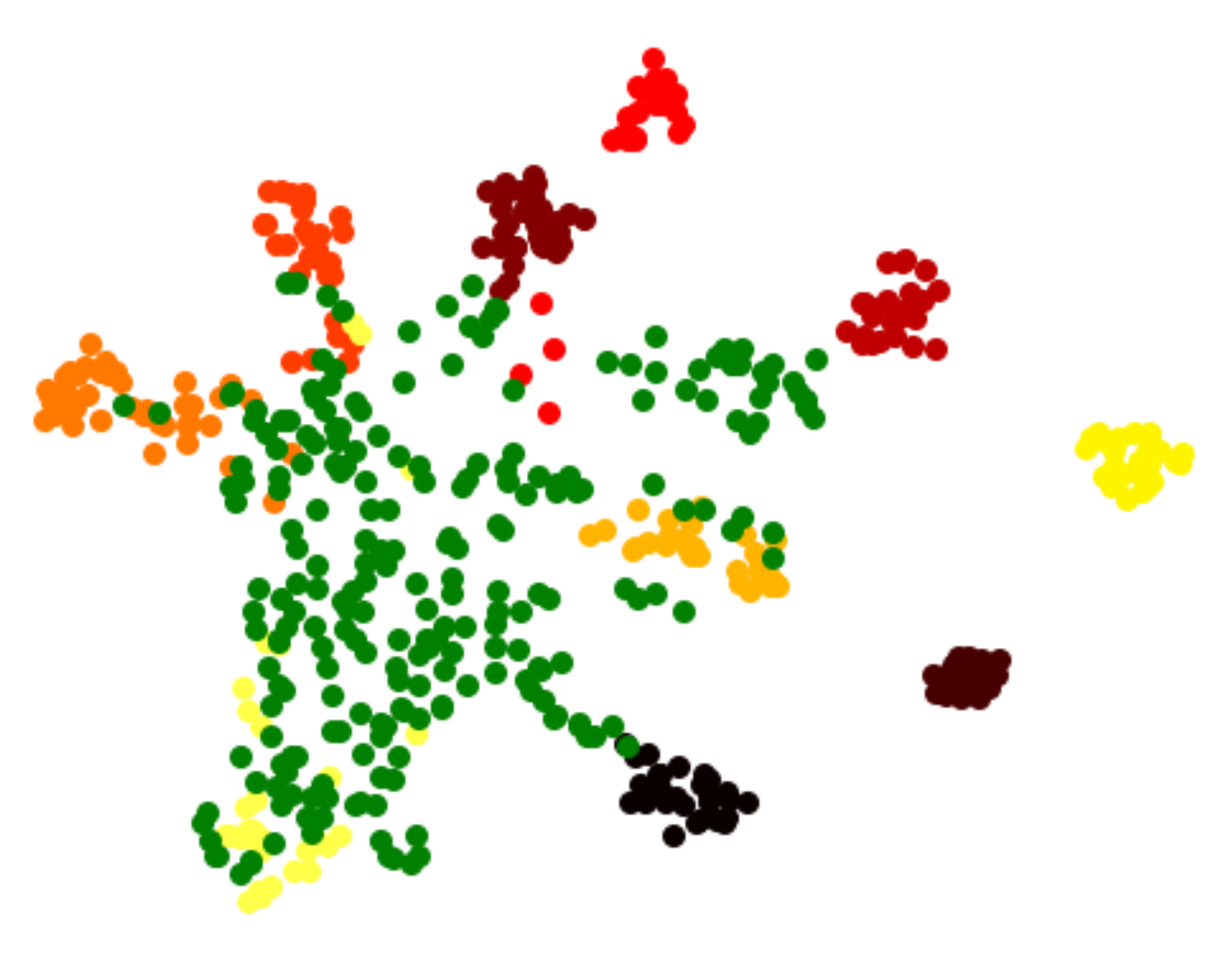}\label{fig:no_adapt}}
 \subfigure[MMD]{
  \includegraphics[width=0.17\hsize]{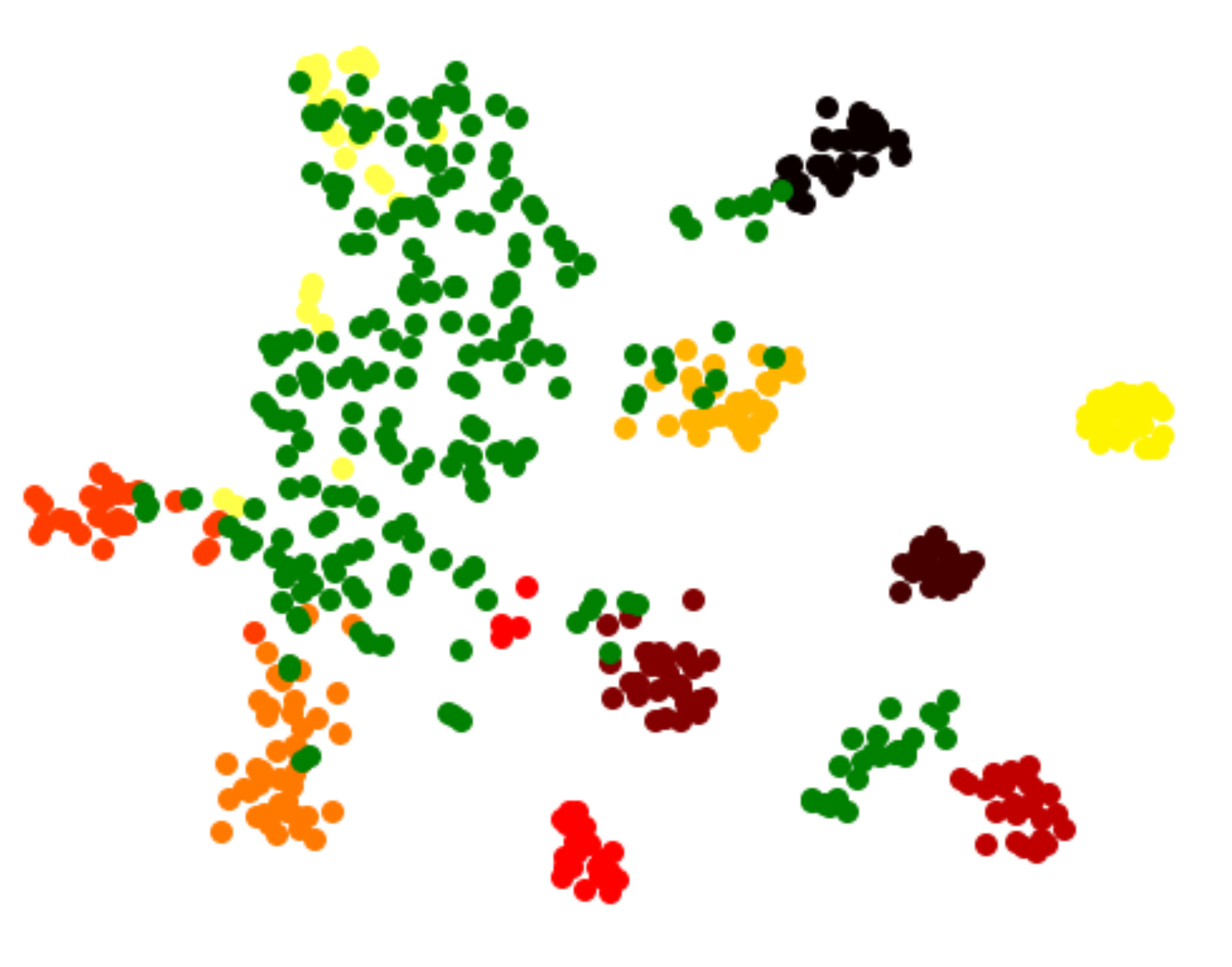}\label{fig:adapt_mmd}}
   \subfigure[BP]{
    \includegraphics[width=0.17\hsize]{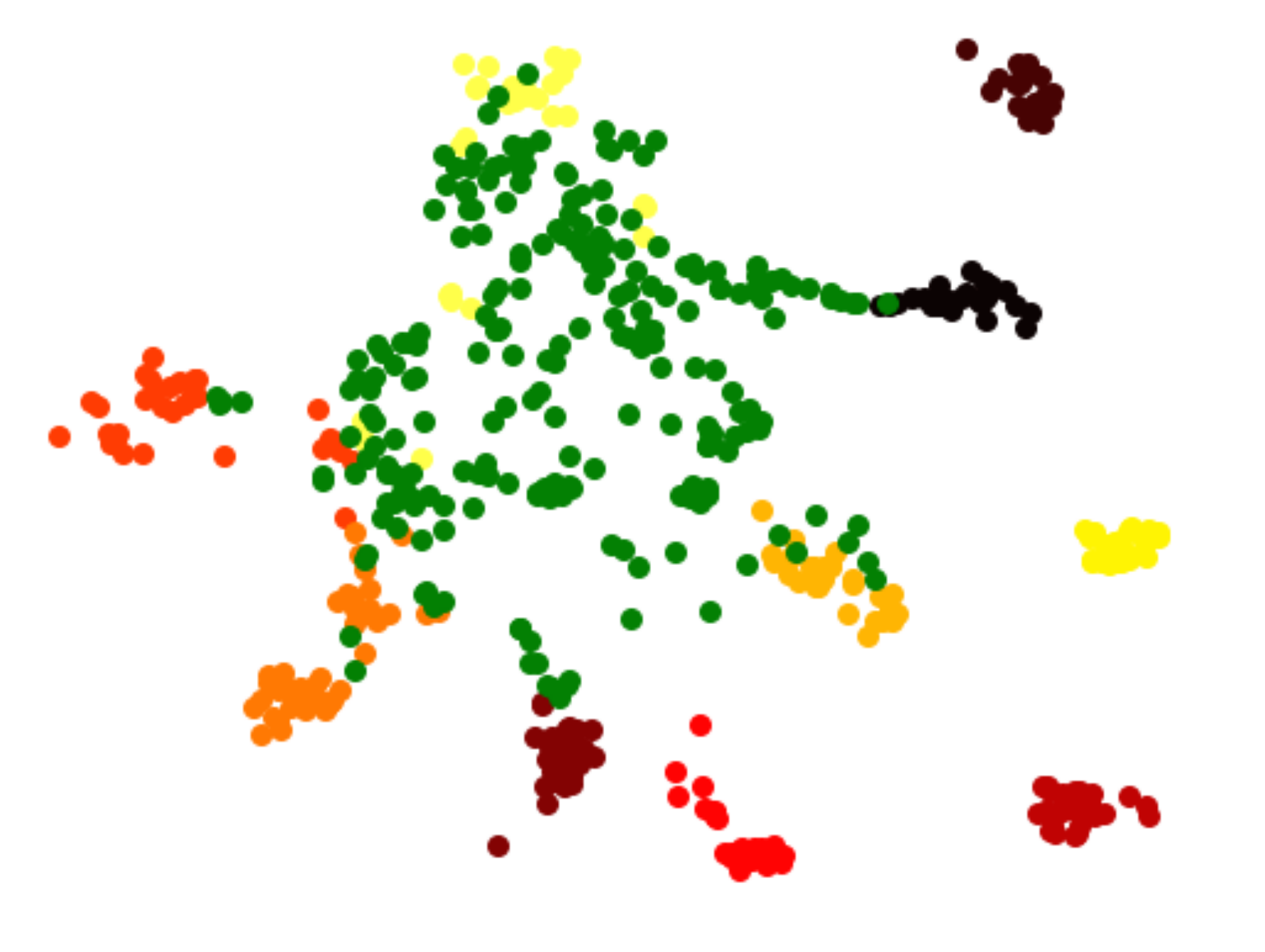}\label{fig:adapt_bp}}
 \subfigure[Ours]{
  \includegraphics[width=0.17\hsize]{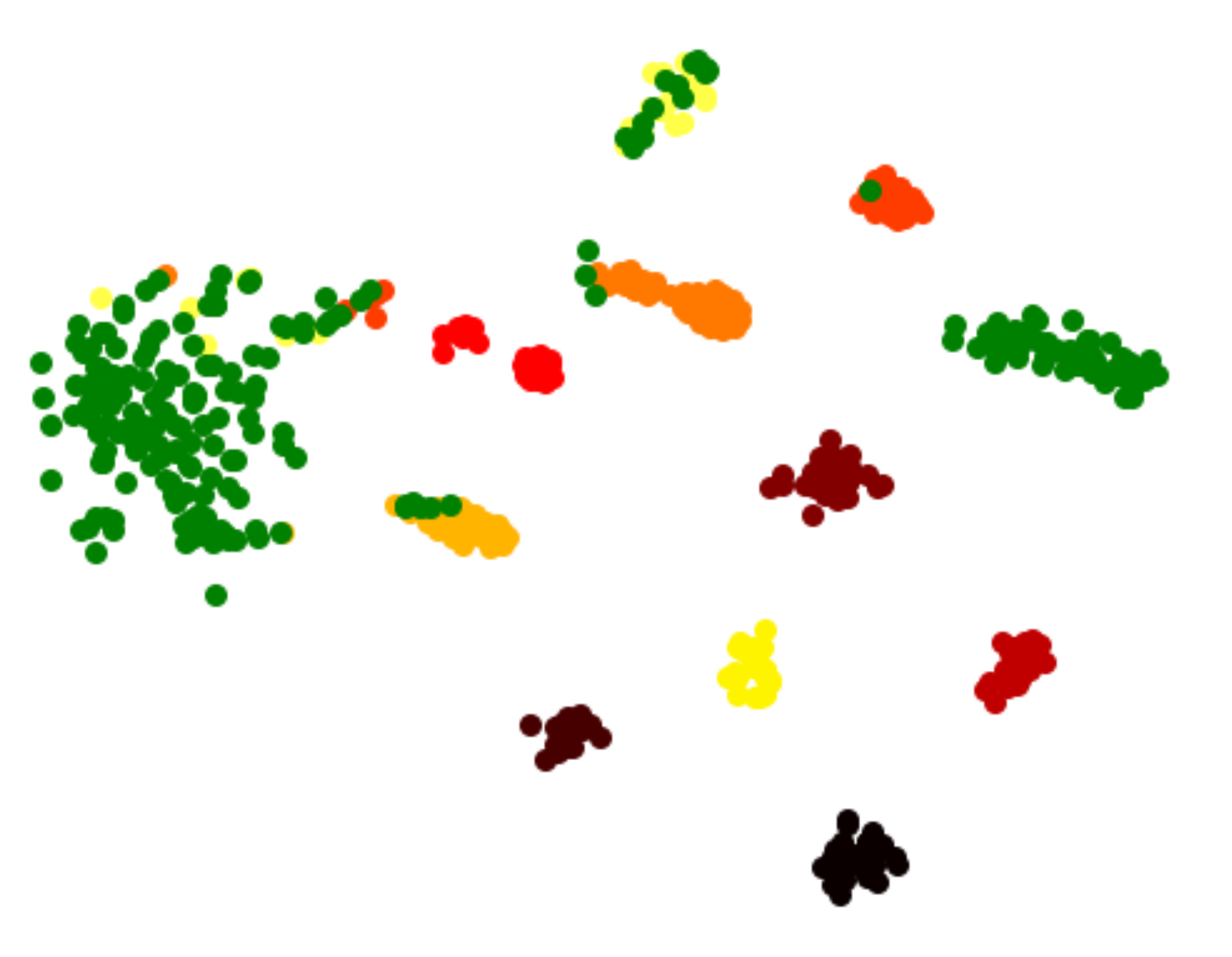}\label{fig:adapt_ours}}
 \vspace{-3mm}
  \caption{Visualization of obtained target features. Green points indicate unknown target samples and points of different colors indicate different classes of target samples. \subref{fig:before_adapt}: Features obtained by  pre-trained AlexNet. \subref{fig:no_adapt}: Features obtained by a model trained without any adaptation. \subref{fig:adapt_mmd}: Features obtained by a model trained with MMD. \subref{fig:adapt_bp}: Features obtained by a model trained with BP. \subref{fig:adapt_ours}: Features obtained by our proposed method. Our method separates unknown target samples from known target samples.}
   \label{vis:feature_vis}
\end{figure}

The results are shown in Table \ref{office:10}. Compared with the baseline methods, our method exhibits better performance in almost all scenarios. The accuracy of the OS is almost always better than that of OS*, which means that many known target samples are regarded as unknown. This is because OSVM is trained to detect outliers and is likely to classify target samples as unknown.
When comparing the performance of OSVM and MMD+OSVM, we can see that the usage of MMD does not always boost the performance. The existence of unknown target samples seems to perturb the correct feature alignment. 
We visualized the learned features using t-SNE~\cite{maaten2008visualizing} in Fig.~\ref{vis:feature_vis}. By training the networks with the source samples, the features seem to be discriminative for known classes. We can see that our method attempts to separate unknown target samples from known ones whereas MMD seems to match most of the target samples with source ones. 
\begin{table}[t]
\centering
\scalebox{0.9}{
\begin{tabular}{c||cc|cc|cc|cc|cc|cc|cc}
\cline{2-15}
 \multicolumn{1}{c}{}  & \multicolumn{14}{|c|}{{\bf Adaptation Scenario}} \\
\cline{2-15}
  \multicolumn{1}{c}{}  & \multicolumn{2}{|c}{A-D}   &  \multicolumn{2}{|c|}{A-W} & \multicolumn{2}{c}{D-A}& \multicolumn{2}{|c|}{D-W}&\multicolumn{2}{c}{W-A}&\multicolumn{2}{|c|}{W-D}&\multicolumn{2}{c|}{AVG} \\
  \multicolumn{1}{c|}{}    &   OS   & OS*    & OS   & OS*  & OS   & OS*   & OS  & OS*& OS  & OS*& OS  & OS*& OS  &  \multicolumn{1}{c|}{OS*} \\\cline{2-7}\cline{2-15}
  \toprule[1.0pt]
   \multicolumn{11}{c}{{\bf Method w/ unknown classes in source domain (AlexNet)}}&\multicolumn{4}{c}{}\\\hline
BP~\cite{ganin2014unsupervised}&78.3&77.3&75.9&73.8&57.6&54.1&89.8&88.9&64.0&61.8&98.7&98.0&77.4&75.7\\
ATI-$\lambda$~\cite{busto2017open}&79.8&79.2&77.6&76.5&71.3&70.0&93.5&93.2&76.7&76.5&98.3&99.2&82.9&82.4\\\hline
\multicolumn{11}{c}{{\bf Method w/o unknown classes in source domain (AlexNet)}}&\multicolumn{4}{c}{}\\\hline
OSVM &59.6&59.1&57.1&55.0&14.3&5.9&44.1&39.3&13.0&4.5&62.5&59.2&40.6&37.1\\
MMD + OSVM&47.8&44.3&41.5&36.2&9.9&0.9&34.4&28.4&11.5&2.7&62.0&58.5&34.5&28.5\\
BP+OSVM&40.8&35.6&31.0&24.3&10.4&1.5&33.6&27.3&11.5&2.7&49.7&44.8&29.5&22.7\\
ATI-$\lambda$\cite{busto2017open} + OSVM&72.0&-&65.3&-&{\bf66.4}&-&82.2&-&71.6&-&92.7&-&75.0&-\\
Ours &  {\bf76.6}	&{\bf 76.4}  & {\bf70.1}&{\bf69.1} &62.5&{\bf 62.3}&{\bf 94.4}&{\bf 94.6}&{\bf 82.3}&{\bf 82.2}&{\bf 96.8}&{\bf 96.9}&{\bf 80.4}&{\bf 80.2}\\\hline
 \multicolumn{11}{c}{{\bf Method w/o unknown classes in source domain (VGGNet)}}&\multicolumn{4}{c}{}\\\hline
OSVM &82.1&83.9&75.9&75.8&38.0&33.1&57.8&54.4&54.5&50.7&83.6&83.3&65.3&63.5\\
MMD + OSVM&84.4&{\bf 85.8}&75.6&75.7&41.3&35.9&61.9&58.7&50.1&45.6&84.3&83.4&66.3&64.2\\
BP+OSVM&83.1&84.7&76.3&76.1&41.6&36.5&61.1&57.7&53.7&49.9&82.9&82.0&66.4&64.5\\
Ours  & {\bf85.8} & {\bf 85.8}    & {\bf 76.9} & {\bf 76.6}& {\bf89.4} &{\bf 91.5} &{\bf 96.0}	&{\bf 96.6}&{\bf 83.4}&{\bf 83.1}&{\bf 97.1}&{\bf97.3}&{\bf 88.0}&{\bf 88.5}\\
\bottomrule[1.0pt]

\end{tabular}}
\caption{Accuracy (\%) of each method in 10 shared class situation. A, D and W correspond to Amazon, DSLR and Webcam respectively. }
\vspace{-5mm}
\label{office:10}
\end{table}

\begin{figure}[t]
\subfigure[\scriptsize{Ratio of unknown}]{
   \includegraphics[width=0.22\hsize]{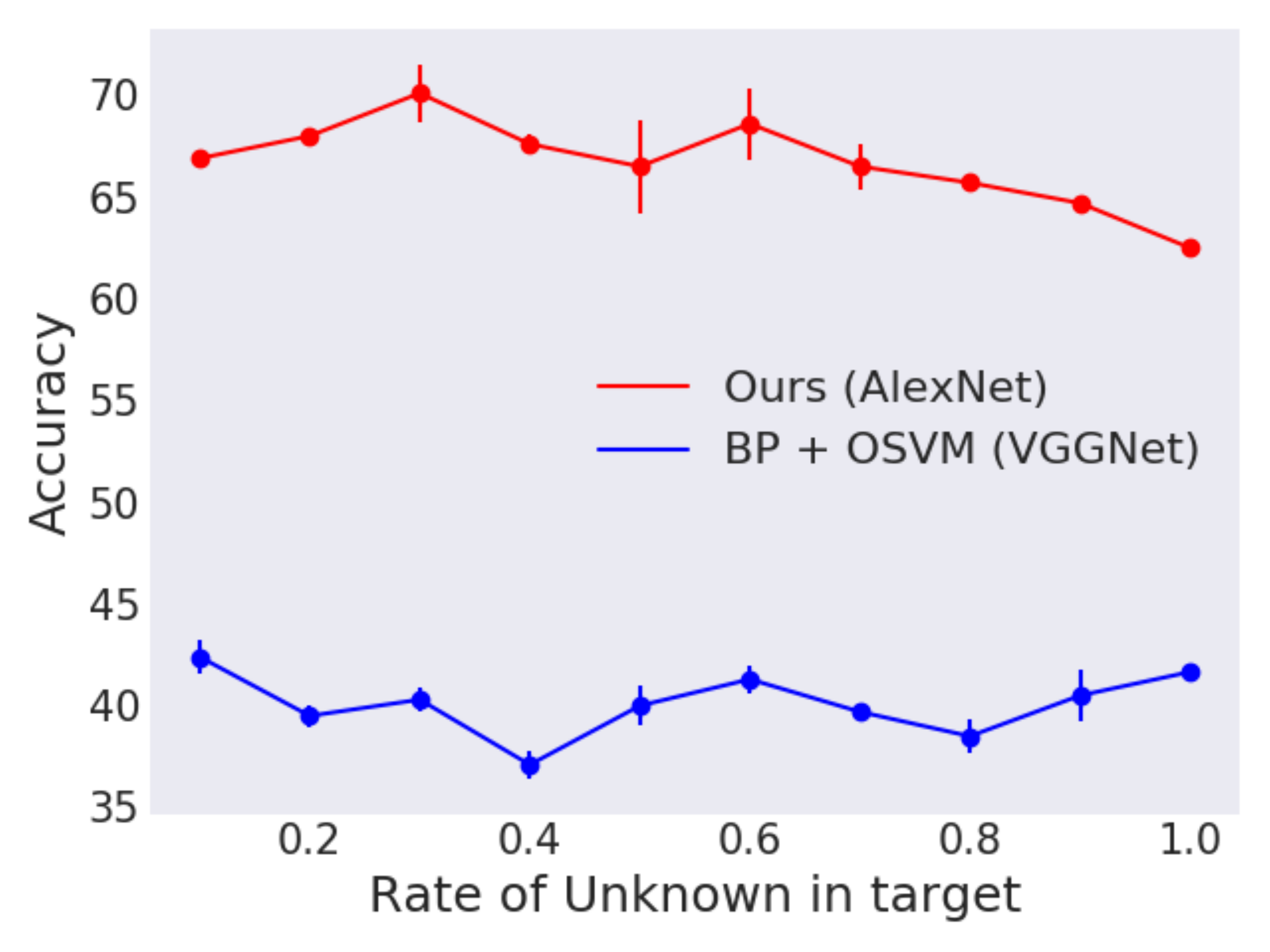}\label{fig:ratio_change}}
   \subfigure[$t$ and accuracy]{
   \includegraphics[width=0.22\hsize]{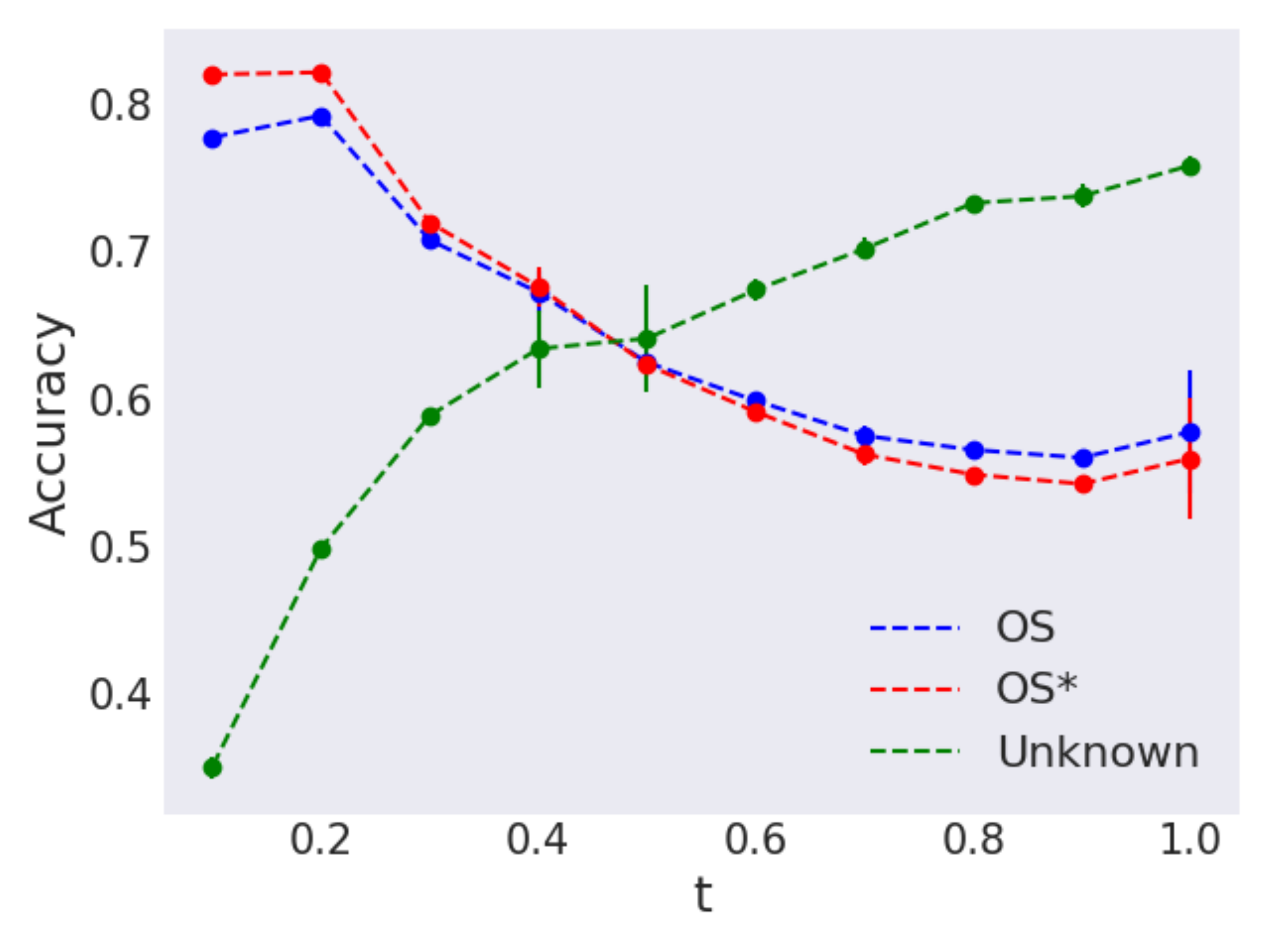}\label{fig:lambd_change}}
  \subfigure[Epoch 50]{
   \includegraphics[width=0.22\hsize]{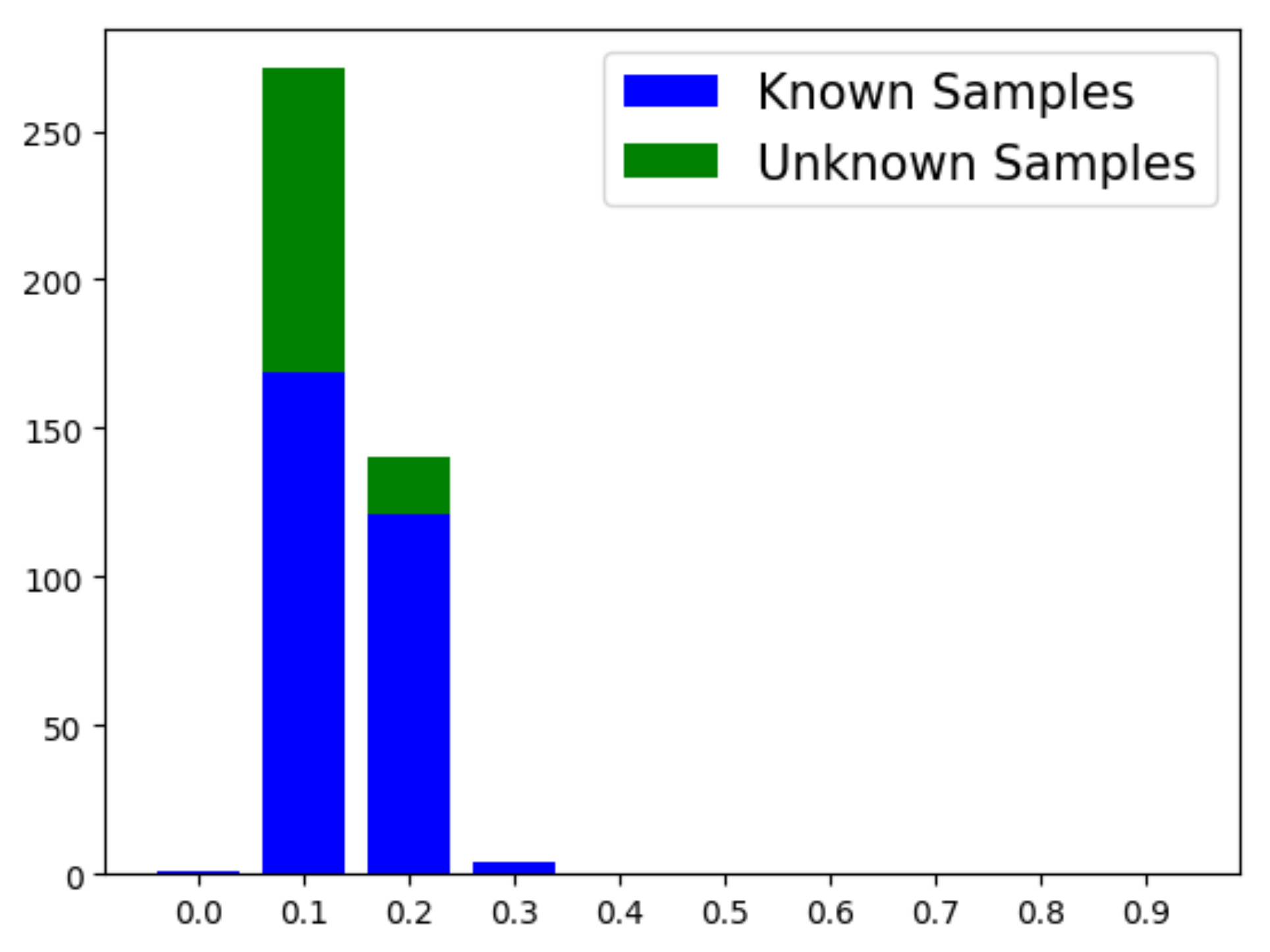}\label{fig:epoch_50}}
   \subfigure[Epoch 500]{
   \includegraphics[width=0.22\hsize]{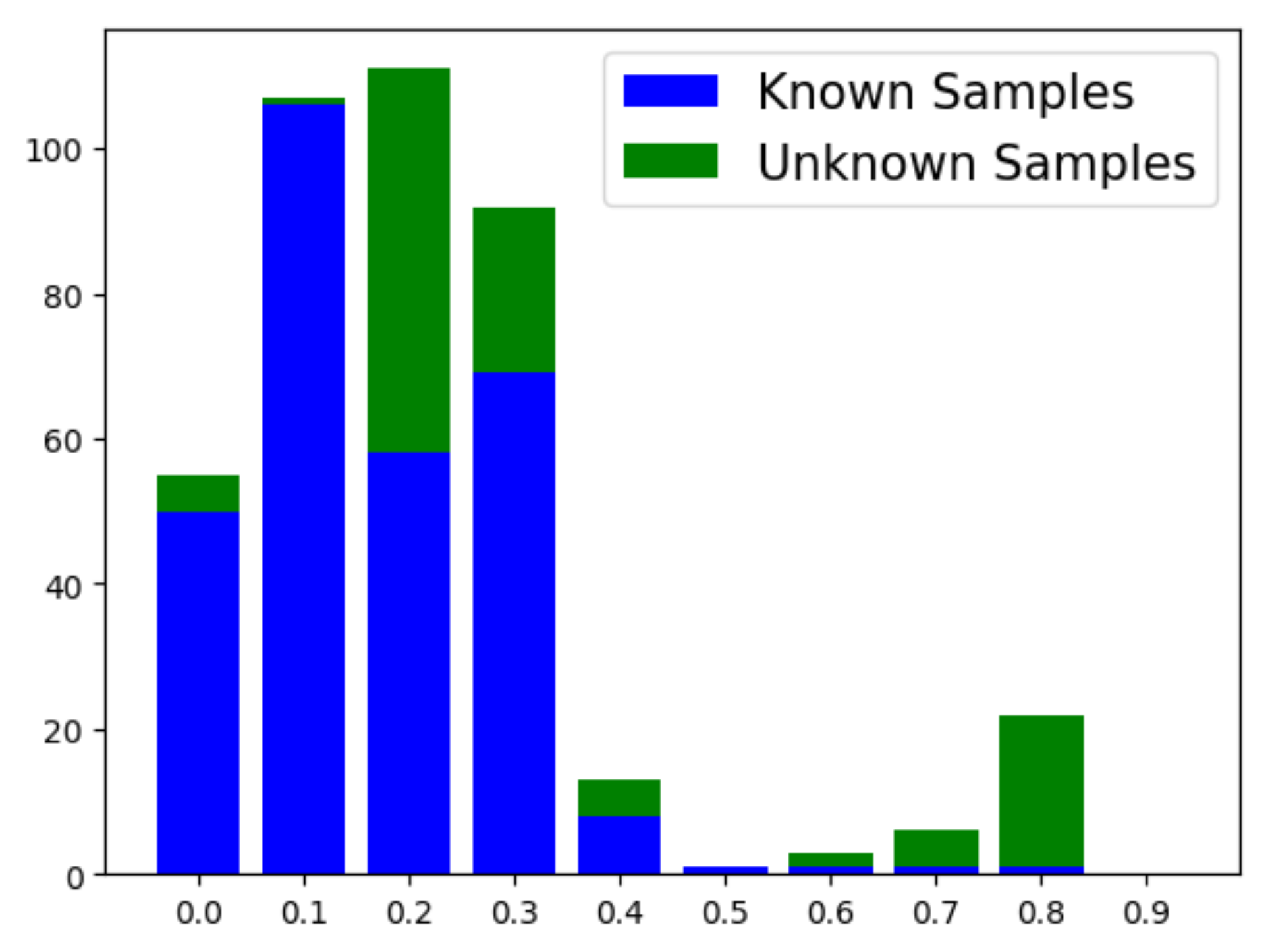}\label{fig:epoch_500}}
   \vspace{-3mm}
  \caption{{\bf \subref{fig:ratio_change}}: The behavior of our method when we changed the ratio of unknown samples in the adaptation from DSLR to Amazon. As we increase the number of unknown target samples, the accuracy decreases. {\bf \subref{fig:lambd_change}}: The change of accuracy with the change of the value $t$ in the same adaptation setting. The accuracy for unknown target samples is denoted as green line. As $t$ increases, target samples are likely classified as "unknown". However, the entire accuracy OS and OS* decrease. {\bf \subref{fig:epoch_50}\subref{fig:epoch_500}}: Frequency diagram of the probability of target samples for unknown class in adaptation from Webcam to DSLR.} 
  \label{fig:analyze}
  \vspace{-3mm}
\end{figure}

\textbf{Number of Unknown Samples and Accuracy}
We further investigate the accuracy when the number of target samples varies in the adaptation from DSLR to Amazon. We randomly chose unknown target samples from Amazon and varied the ratio of the unknown samples. The accuracy of OS is shown in Fig. \ref{fig:ratio_change}. When the ratio changes, our method seems to perform well. 
\textbf{Value of ${\mathbf t}$}
We observe the behavior of our model when the training signal, $t$ in Eq. \ref{eq:loss_adv} is varied. As we mentioned in the method section, When $t$ is equal to $1$, the objective of the generator is to match the whole distribution of the target features with that of the source, which is exactly the same as an existing distribution matching method. Accordingly, the accuracy should degrade in this case. According to Fig. \ref{fig:analyze}\subref{fig:lambd_change}, as we increase the value of $t$, the accuracies of OS and OS* decrease and the overall accuracy increases. This result means that the model does not learn representations where unknown samples from known samples can be distinguished. 

\textbf{Probability for Unknown Class}
In Fig. \ref{fig:analyze}\subref{fig:epoch_50}\subref{fig:epoch_500}, frequency diagram of the probability for an unknown class is shown in the adaptation from Webcam to DSLR dataset. At the beginning of training, Fig. \ref{fig:analyze}\subref{fig:epoch_50}, the probability is low in most samples including the known and unknown samples. As shown in Fig. \ref{fig:analyze}\subref{fig:epoch_500}, many unknown samples have high probability for unknown class whereas many known samples have low probability for the class after training the model for 500 epochs. We can observe that unknown and known samples seem to be separated from the result.
\vspace{-5mm}
\subsubsection{21 Class Classification}\label{21class}

In addition, we observe the behavior of our method when the number of known classes increases. We add the samples of 10 classes which were not used in the previous setting. The 10 classes are the ones used as unknown samples in the source domain in~\cite{busto2017open}. In total, we conducted 21 class classification experiments in this setting. We also evaluate our method on VGG Network. With regard to other details of the experiment, we followed the setting of the previous experiment. 

The results are shown in Table~\ref{office_20}. Compared to the baseline methods, the superiority of our method is clear. The usefulness of MMD and BP is not observed for this setting too. 
An examination of the result of adaptation from Amazon to Webcam (A-W) reveals that the accuracy of other methods is better than our approach based on OS* and OS. However, \doublequote{ALL} of the measurements are inferior to our method. The value of \doublequote{ALL} indicates the accuracy measured for all the samples without averaging over classes. Thus, the result means that existing methods are likely to recognize target samples as one of known classes in this setting. 
From the results, the effectiveness of our method is verified when the number of class increases.

\begin{table}[t]
\centering
\scalebox{0.75}{
\begin{tabular}{l||ccc|ccc|ccc|ccc}
\cline{2-10}
\multicolumn{1}{c}{}  & \multicolumn{9}{|c|}{{\bf Adaptation Scenario}} \\\cline{2-10}
  \multicolumn{1}{c}{}  & \multicolumn{3}{|c}{A-D}   &  \multicolumn{3}{|c|}{A-W} & \multicolumn{3}{c|}{D-A}&\multicolumn{3}{c}{}\\
  \multicolumn{1}{c|}{}    &   OS   & OS*  &ALL & OS   & OS* &ALL & OS   & OS*   &ALL\\\cline{1-10}\cline{1-10}\cline{1-10}
OSVM &73.6$\pm$0.4&{\bf 75.8}$\pm$0.6&57.6&{\bf 72.0}$\pm$0.5&{\bf 74.1}$\pm$0.5&58.0&44.9$\pm$0.1&	43.9$\pm$0.1&51.1\\
MMD + OSVM &72.1$\pm$0.9&73.9$\pm$1.0&57.8&69.1$\pm$0.8&71.2$\pm$0.9&54.9&29.8$\pm$0.6&26.5$\pm$0.6&50.3\\
BP + OSVM &70.4$\pm$0.2&72.1$\pm$0.3&57.1&70.9$\pm$0.5&72.9$\pm$0.4&57.6&30.9$\pm$0.2&27.6$\pm$0.2&51.3\\
Ours & {\bf74.8}$\pm$0.5 &74.6 $\pm$0.5&{\bf 73.9}& 66.8$\pm$3.5&66.1$\pm$3.7&{\bf 69.7}& {\bf 64.6}$\pm$1.2	&{\bf 65.9}$\pm$4.9&{\bf 68.5}\\\hline\hline
\multicolumn{1}{c}{}& \multicolumn{3}{|c|}{D-W}&\multicolumn{3}{c}{W-A}&\multicolumn{3}{|c|}{W-D}&\multicolumn{3}{c}{AVG}\\
\multicolumn{1}{c|}{} & OS  & OS* &ALL& OS  & OS* &ALL& OS  & OS* &ALL& OS  & OS*  &ALL\\\hline
OSVM &63.1$\pm$1.1&61.9$\pm$1.2&69.9&34.0$\pm$0.9&31.8$\pm$1.3&48.3&82.9$\pm$2.3&82.9$\pm$1.7&84.2&61.8&61.7&61.5\\
MMD + OSVM &58.3$\pm$0.6&56.6$\pm$0.6&68.8&39.7$\pm$2.1&37.1$\pm$2.4&55.9&84.5$\pm$1.2&84.2$\pm$1.3&87.2&58.9&58.2&62.3\\
BP+OSVM&63.2$\pm$2.8&61.7$\pm$3.0&71.3&40.0$\pm$2.7&37.4$\pm$3.0&56.0&83.5$\pm$0.8&83.1$\pm$0.8&86.4&59.8&59.1&63.2\\
Ours&{\bf 83.1}$\pm$0.6&{\bf 82.5}$\pm$0.6&{\bf 84.9}&{\bf 65.9}$\pm$0.1&{\bf65.3}$\pm$0.2&{\bf 69.0}&{\bf 92.8}$\pm$0.2&{\bf 93.3}$\pm$0.2&{\bf 90.3}&{\bf 74.7}&{\bf 74.6}&{\bf 76.1}\\
\bottomrule[1.0pt]
\end{tabular}}
\caption{Accuracy (\%) of experiments on Office dataset in 20 shared class situation. We used VGG Network to obtain the results.}
\vspace{-3mm}
\label{office_20}
\end{table}
\vspace{-3mm}
\subsection{Experiments on VisDA Dataset}
\vspace{-3mm}
We further evaluate our method on adaptation from synthetic images to real images. 
VisDA dataset~\cite{visda} consists of 12 categories in total. The source domain images are collected by rendering 3D models whereas the target domain images consist of real images. We used the training split as the source domain and validation one as the target domain. 
We choose 6 categories (bicycle, bus, car, motorcycle, train and truck) from them and set other 6 categories as the unknown class (aeroplane, horse, knife, person, plant and skateboard). The training procedure of the networks is the same as that used for Office dataset. 

The results are shown in Table \ref{visda_result}. Our method outperformed the other methods in most classes and on average. {\it Avg} indicates the accuracy averaged over all classes. {\it Avg known} indicates the accuracy averaged over only known classes. In both evaluation metrics, our method showed better performance, which means that our method is better both at matching distributions between known samples and rejecting unknown samples in open set domain adaptation setting.
In this setting, the known classes and unknown class should have different characteristics because known classes are picked up from vehicles and unknown samples are from others. Thus, in our method, the accuracy for the unknown class is better than that for the known classes. 
We further show the examples of images in Table \ref{vis:visda_ex}. Some of the known samples are recognized as unknown. As we can see from the three images, most of them contain multiple classes of objects or are hidden by other objects. Then, look at the second columns from the left. The images are categorized as motorcycle though they are unknown. The images of motorcycle often contain persons and the appearance of the person and horse have similar features to such images. In the third and fourth columns, we demonstrate the correctly classified known and unknown samples. If the most part of the image is occupied by the object of interest, the classification seems to be successful. 
\begin{table}[t]
\centering
\scalebox{0.9}{
\begin{tabular}{l||ccccccc||cc}
\toprule[1.5pt]
{\bf Method}&\rotatebox{60}{bcycle}
 & \rotatebox{60}{bus}  & \rotatebox{60}{car}  & \rotatebox{60}{mcycle} & \rotatebox{60}{train}& \rotatebox{60}{truck} & \rotatebox{60}{unknwn} & \rotatebox{60}{Avg} & \rotatebox{60}{Avg knwn}  \\\hline
\multicolumn{1}{c}{{\bf AlexNet}}&\multicolumn{9}{c}{}\\\hline
OSVM&4.8&45.0&44.2&43.5&59.0&10.5&57.4&37.8&34.5\\
OSVM+MMD&0.2&30.9&49.1&54.8&56.1&8.1&61.3&37.2&33.2\\
OSVM+BP&9.1&50.5&{\bf 53.9}&79.8&69.0&8.1&42.5&44.7&45.1\\\hline
Ours &{\bf 48.0} &{\bf 67.4} & 39.2& {\bf 80.2}  & {\bf 69.4}  & {\bf 24.9}&{\bf 80.3} &{\bf 58.5}&{\bf 54.8} \\\hline
\multicolumn{1}{c}{{\bf VGGNet}}&\multicolumn{9}{c}{}\\\hline
OSVM&31.7&51.6&66.5&70.4&{\bf 88.5}&20.8&38.0&52.5&54.9\\
OSVM+MMD&39.0&50.1&64.2&79.9&86.6&16.3&44.8&54.4&56.0\\
OSVM+BP&31.8&56.6&{\bf 71.7}&77.4&87.0&22.3&41.9&55.5&57.8\\\hline
Ours & {\bf 53.6} & {\bf 72.0} & 49.1& {\bf 80.8}& 81.9 &{\bf  29.4}  &{\bf 89.7} & {\bf 65.2}&{\bf 61.1}\\
\bottomrule[1.5pt]
\end{tabular}}
\caption{Accuracy (\%) on VisDA dataset. The accuracy per class is shown. }
\label{visda_result}
\vspace{-3mm}
\end{table}
\begin{table}[h]
\centering
\scalebox{0.8}{
  \begin{tabular}{c|c|c|c}
  \toprule [1.0pt]
  \multicolumn{4}{c}{{\bf Ground Truth Class $\rightarrow$ Predicted Class}}\\
    \hline \scriptsize {\bf Known $\rightarrow$ Unknown $\times$}& \scriptsize {\bf Unknown  $\rightarrow$ Known $\times$}&\scriptsize {\bf Known $\rightarrow$ Known $\surd$}& \scriptsize {\bf Unknown $\rightarrow$ Unknown $\surd$}\\ \hline\hline
    
    \scriptsize Train $\rightarrow$ Unknown & \scriptsize Unknown  $\rightarrow$ Motorcycle &\scriptsize Truck $\rightarrow$ Truck & \scriptsize Unknown $\rightarrow$ Unknown\\
    \begin{minipage}{0.24\hsize}
      \centering
    \includegraphics[width=0.8\hsize,height=0.8\hsize]{figure/train_unk.pdf}
    \end{minipage} &
    \begin{minipage}{0.24\hsize}
      \centering
      \includegraphics[width=0.8\hsize,height=0.8\hsize]{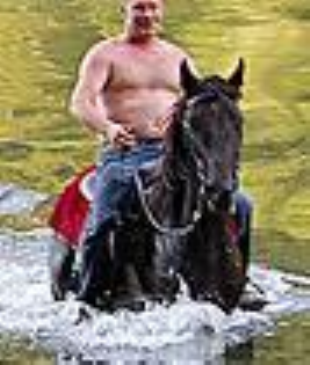}
    \end{minipage} &
    \begin{minipage}{0.24\hsize}
      \centering
    \includegraphics[width=0.8\hsize,height=0.8\hsize]{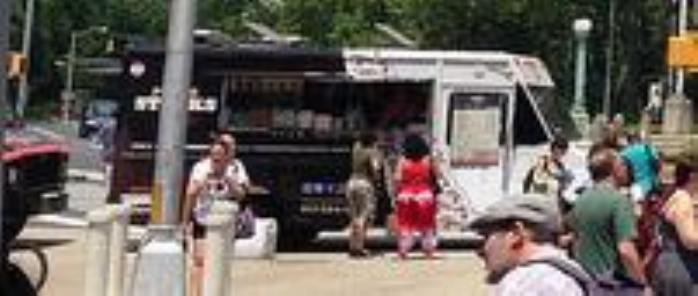}
    \end{minipage} &
    \begin{minipage}{0.24\hsize}
      \centering
      \includegraphics[width=0.8\hsize,height=0.8\hsize]{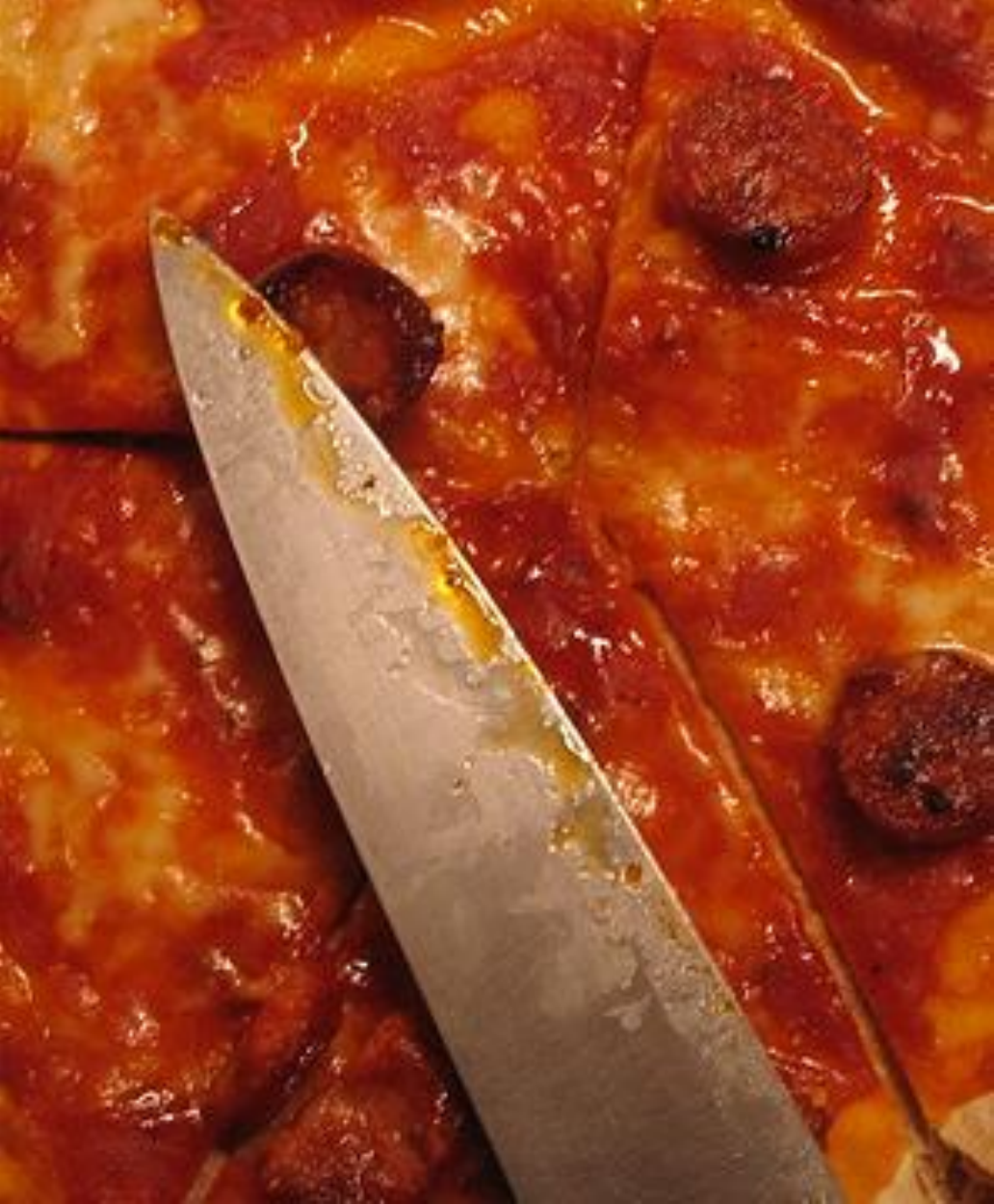}
    \end{minipage} 
    \\\hline
     \scriptsize Motorcycle $\rightarrow$ Unknown & \scriptsize Unknown  $\rightarrow$ Motorcycle &\scriptsize Bicycle $\rightarrow$ Bicycle & \scriptsize Unknown $\rightarrow$ Unknown\\
    \begin{minipage}{0.24\hsize}
      \centering
    \includegraphics[width=0.8\hsize,height=0.8\hsize]{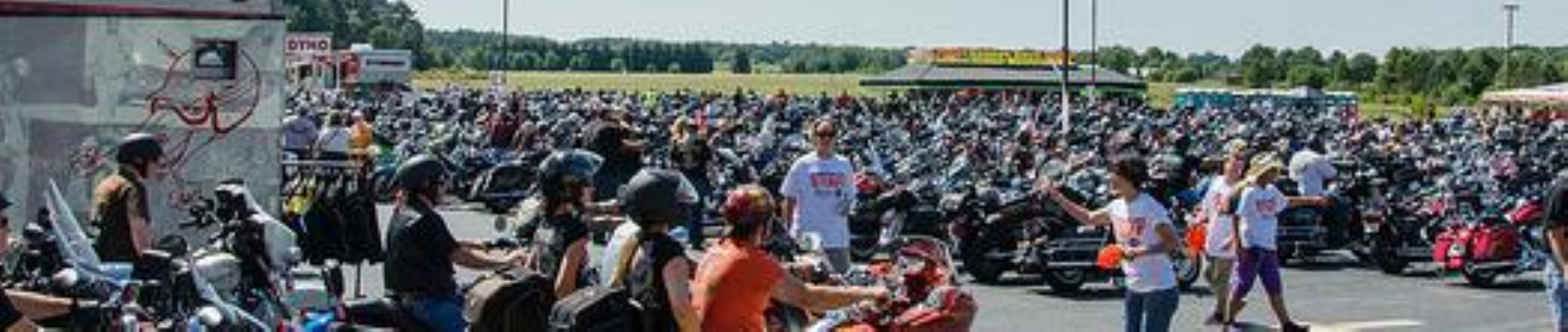}
    \end{minipage} &
    \begin{minipage}{0.24\hsize}
      \centering
      \includegraphics[width=0.8\hsize,height=0.8\hsize]{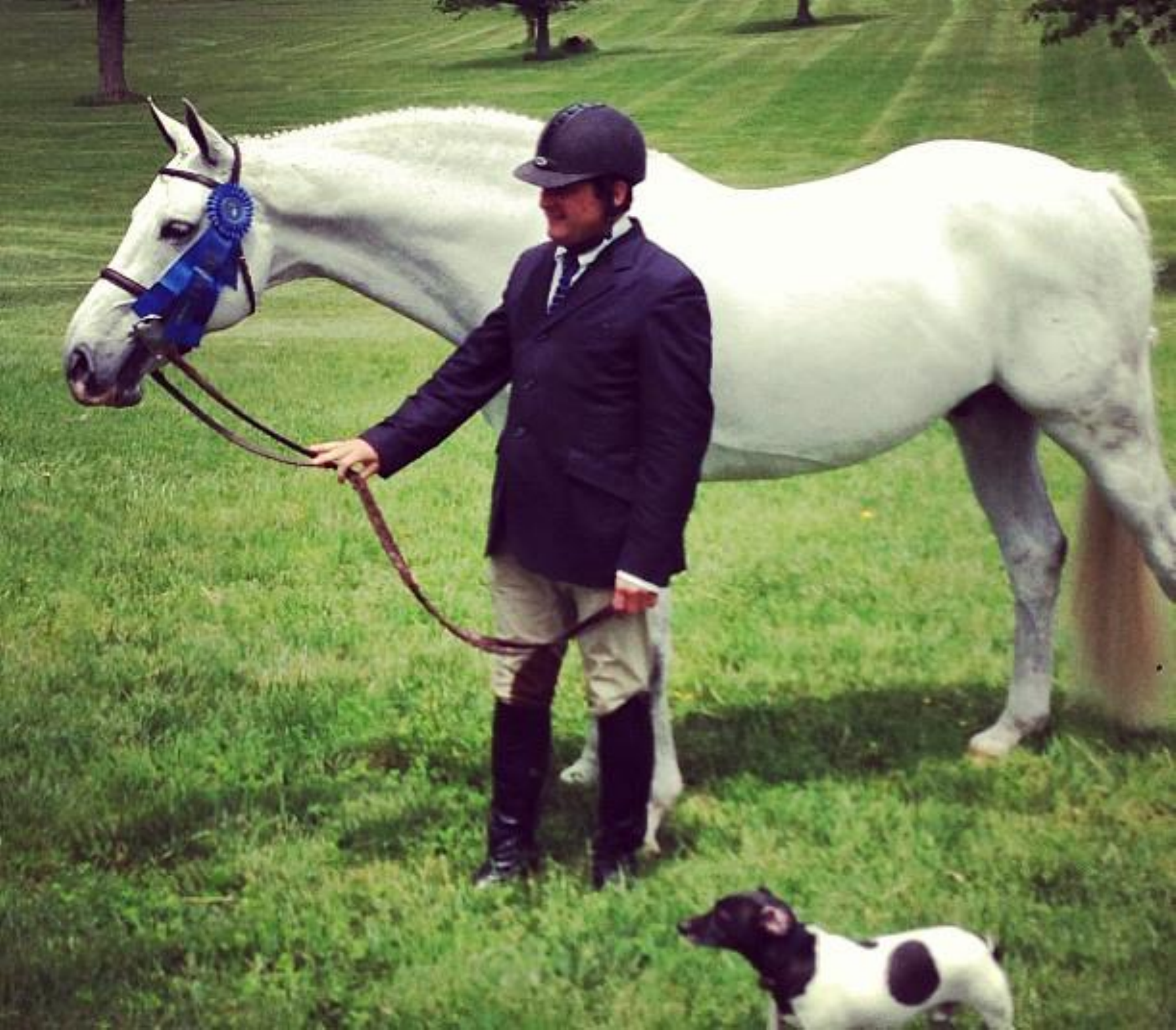}
    \end{minipage} &
    \begin{minipage}{0.24\hsize}
      \centering
    \includegraphics[width=0.8\hsize,height=0.8\hsize]{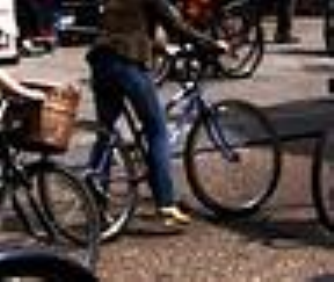}
    \end{minipage} &
    \begin{minipage}{0.24\hsize}
      \centering
      \includegraphics[width=0.8\hsize,height=0.8\hsize]{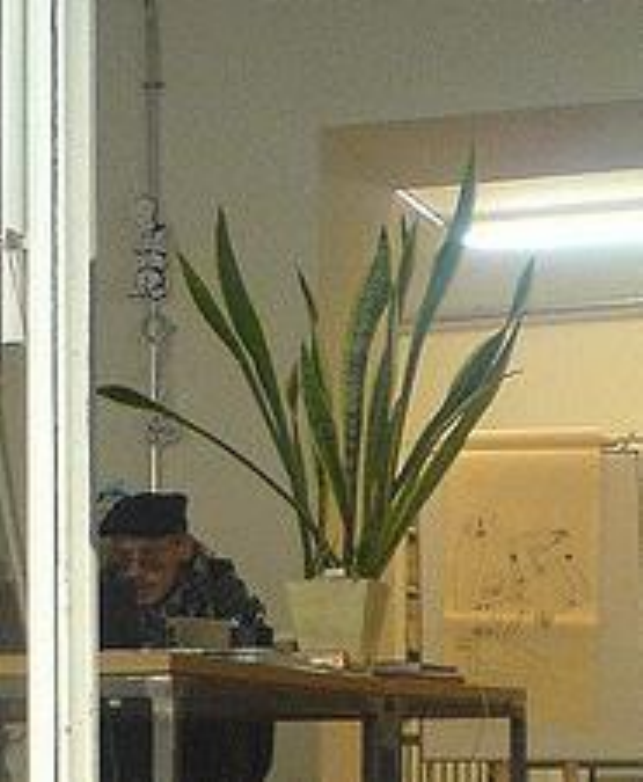}
    \end{minipage} 
    \\ \hline 
    \scriptsize Car $\rightarrow$ Unknown & \scriptsize Unknown  $\rightarrow$ Motorcycle &\scriptsize Motorcycle $\rightarrow$ Motorcycle & \scriptsize Unknown $\rightarrow$ Unknown\\
    \begin{minipage}{0.24\hsize}
      \centering
    \includegraphics[width=0.8\hsize,height=0.8\hsize]{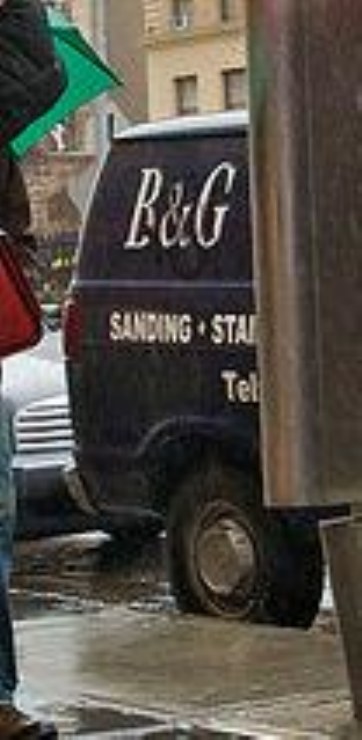}
    \end{minipage} &
    \begin{minipage}{0.24\hsize}
      \centering
      \includegraphics[width=0.8\hsize,height=0.8\hsize]{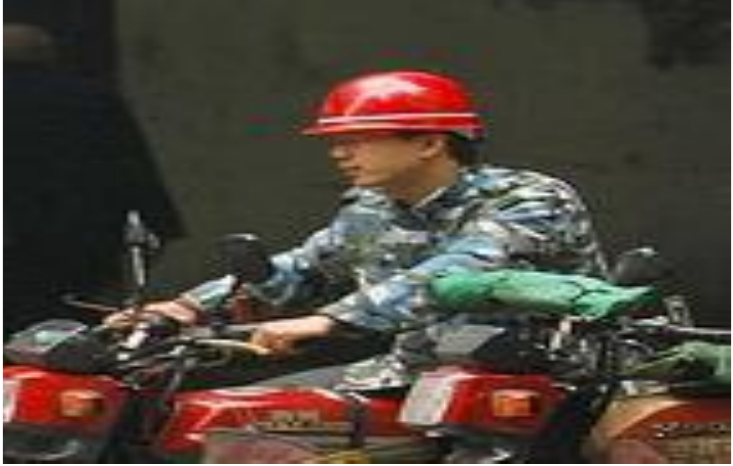}
    \end{minipage} &
    \begin{minipage}{0.24\hsize}
      \centering
    \includegraphics[width=0.8\hsize,height=0.8\hsize]{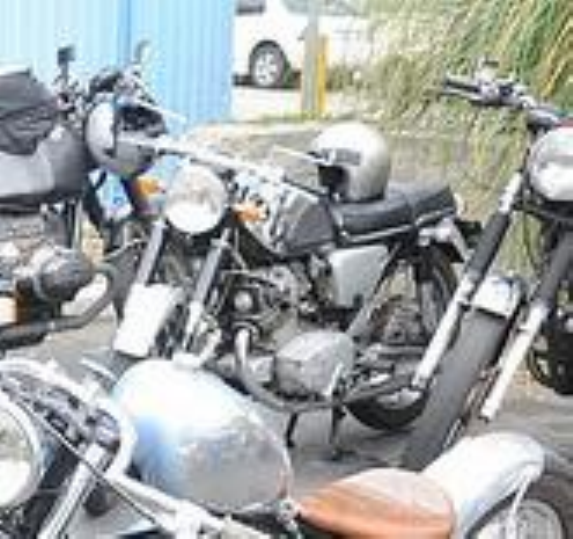}
    \end{minipage} &
    \begin{minipage}{0.24\hsize}
      \centering
      \includegraphics[width=0.8\hsize,height=0.8\hsize]{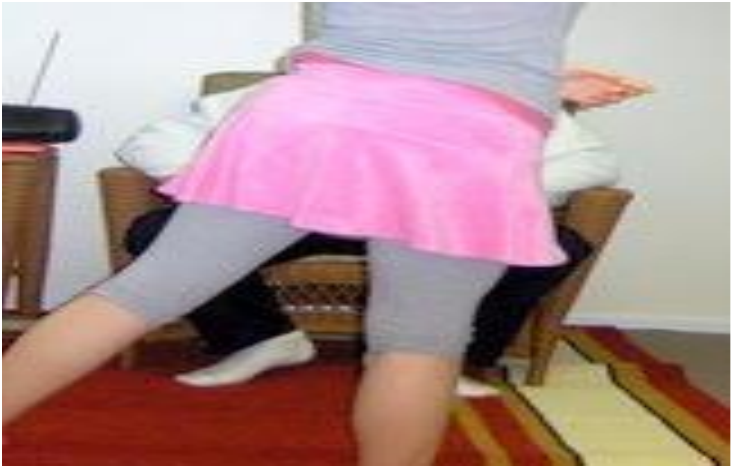}
      \\
    \end{minipage} \\
      \bottomrule [1.0pt]
  \end{tabular}}
  \caption{Examples of recognition results on VisDA dataset.}
 \label{vis:visda_ex}
 \vspace{-7mm}
\end{table}
\vspace{-4mm}
\subsection{Experiments on Digits Dataset}
\vspace{-3mm}
We also evaluate our method on digits dataset. We used SVHN~\cite{netzer2011reading},USPS~\cite{lecun1998gradient} and MNIST for this experiment. In this experiment, we conducted 3 scenarios in total. Namely, adaptation from SVHN to MNIST, USPS to MNIST and MNIST to USPS. These are common scenarios in unsupervised domain adaptation. The numbers from 0 to 4 were set as known categories whereas the other numbers were set as unknown categories. In this experiment, we also compared our method with two baselines, OSVM and MMD combined with OSVM. With regard to OSVM, we first trained the network using source known samples and extracted features using the network, then applied OSVM to the features. When training CNN, we used Adam~\cite{kingma2014adam} with a learning rate $2.0\times10^{-5}$.

\textbf{Adaptation from SVHN to MNIST}
In this experiment, we used all SVHN training samples with numbers in the range from 0 to 4 to train the network. With regard to the target samples, we used all samples in the training splits of MNIST.

\textbf{Adaptation between USPS and MNIST}
When using the datasets as a source domain, we used all training samples with number from 0 to 4. With regard to the target datasets, we used all training samples.

\begin{figure}[t]

\subfigure[Source Only]{
   \includegraphics[width=0.23\hsize]{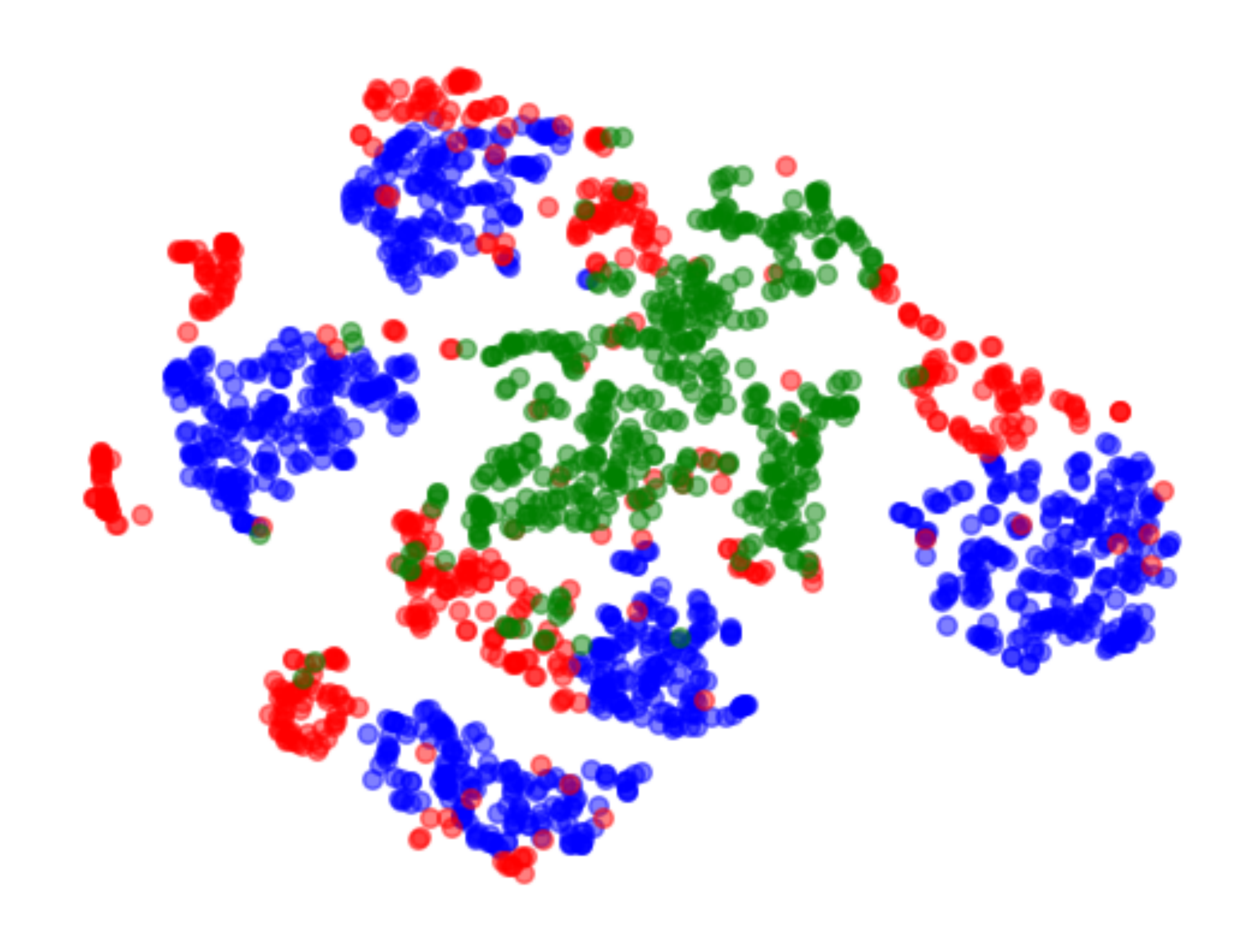}}
  \subfigure[MMD]{
   \includegraphics[width=0.23\hsize]{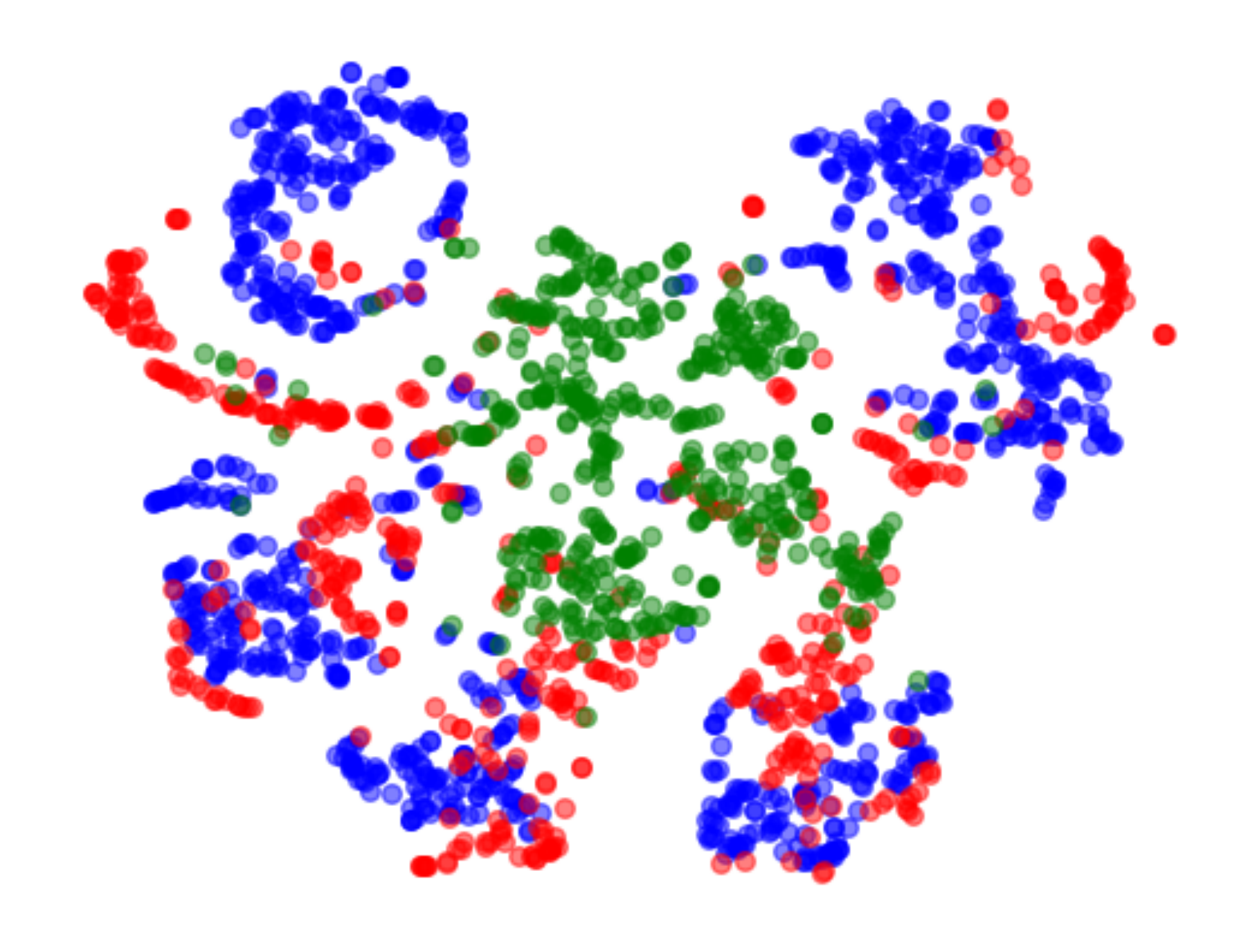}}
  \subfigure[BP]{
   \includegraphics[width=0.23\hsize]{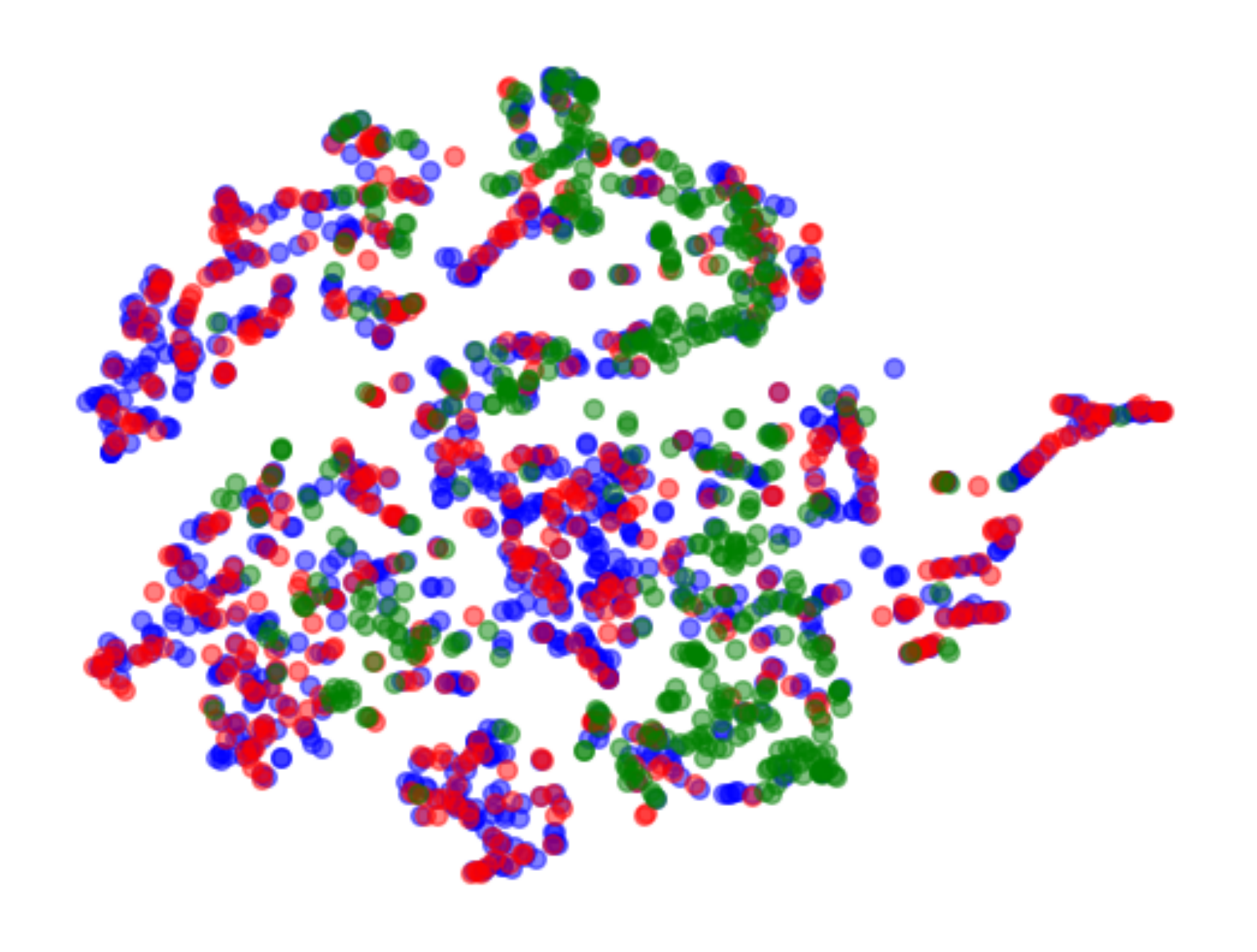}}
  \subfigure[Ours]{
   \includegraphics[width=0.23\hsize]{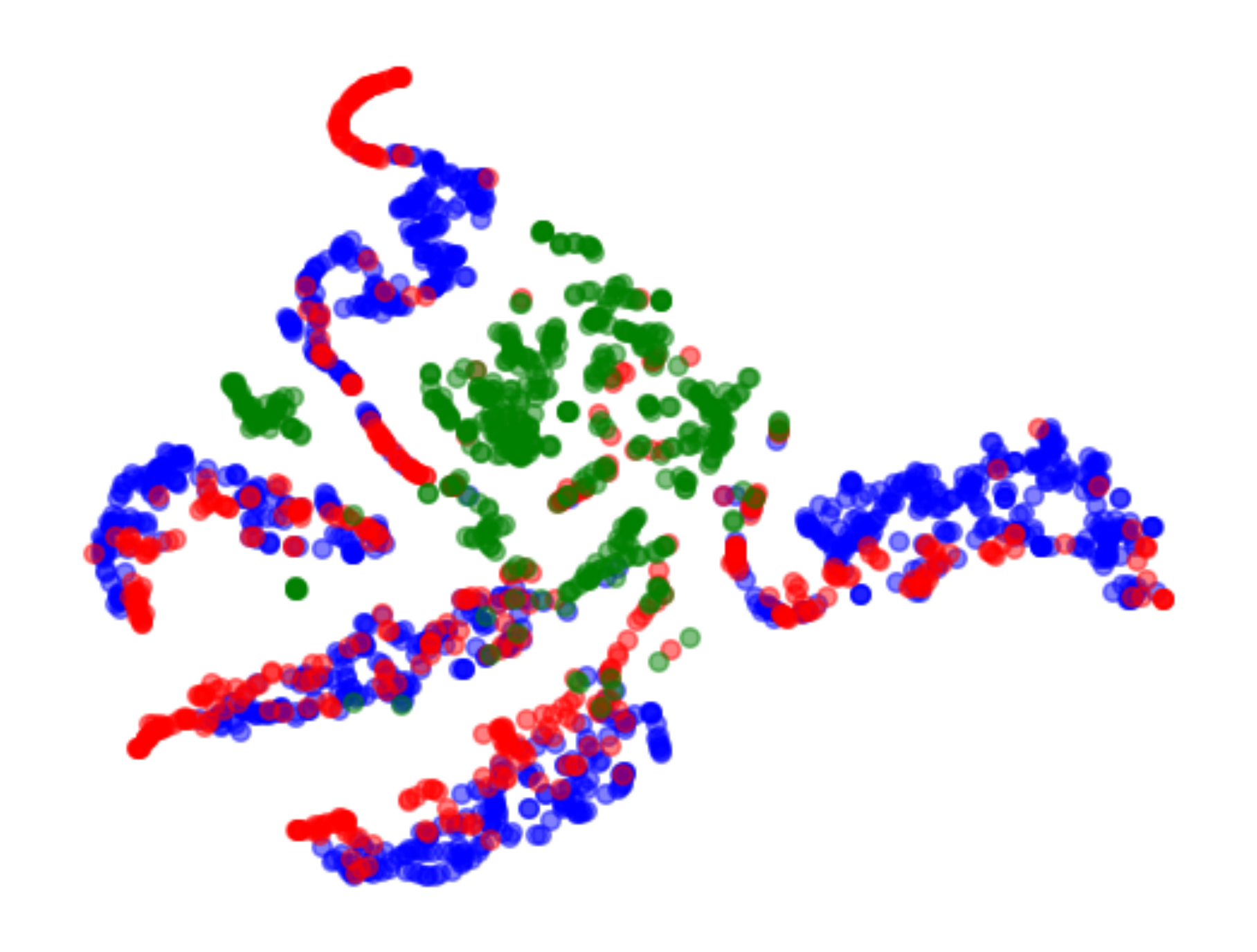}}
\vspace{-3mm}
  \caption{Feature visualization of adaptation from USPS to MNIST. Visualization of source and target features. {\bf Blue points} are source features. {\bf Red points} are target known features. {\bf Green points} are target unknown features.}
   \label{vis:feature_digit}
   
\end{figure}

\textbf{Result}
The qualitative results are shown in Table \ref{exp_digits}. Our proposed method outperformed other methods. In particular, with regard to the adaptation between USPS and MNIST, our method achieves accurate recognition. In contrast, the adaptation performance on for SVHN to MNIST is worse compared to the adaptation between USPS and MNIST. Large domain different between SVHN and MNIST causes the bad performance. 
We also visualized the learned features in Fig. \ref{vis:feature_digit}. Unknown classes (5$\sim$9) are separated using our method whereas known classes are aligned with source samples. The method based on distribution matching such as BP~\cite{ganin2014unsupervised} fails in adaptation for this open set scenario. When examining the learned features, we can observe that BP attempts to match all of the target features with source features. Consequently, unknown target samples are made difficult to detect, which is obvious from the qualitative results for BP. The accuracy of UNK in BP+OSVM is much worse than the other methods.
\vspace{-3mm}
\subsection{Application to Semi-supervised Open Set Recognition}\label{std_opnst}
\vspace{-3mm}
In this section, we conduct experiments for the situation where we are provided with known samples and unlabeled samples. The domain of the known and unlabeled samples is the same. 
In this setting, the numbers from 0 to 4 are regarded as known and the others as unknown. For the labeled samples, we pick 1000 samples per class and rest of the samples are treated as unlabeled samples. We trained each method in the same way as done in the previous experiment. 
The result is demonstrated in Table \ref{op_setting}. OSVM performs well because there is no domain-shift between training and testing samples. In addition, our method mostly performed better than OSVM. With regard to the classification of SVHN, our proposed method seems to fail in classifying unknown samples. However, when combined with OSVM (Ours + OSVM), the accuracy improved dramatically, which indicates that the generator learned good representations to reject unknown samples whereas a good classifier was not trained in this setting. 

\begin{table}[t]
\centering
\scalebox{0.8}{
\begin{tabular}{c||cccc|cccc|cccc||cccc}
 \toprule[1.5pt]
& \multicolumn{4}{|c}{SVHN-MNIST}   &  \multicolumn{4}{|c|}{USPS-MNIST} & \multicolumn{4}{|c||}{MNIST-USPS}&\multicolumn{4}{|c}{Average}\\
  \multicolumn{1}{c||}{Method}    &   OS   & OS*  &ALL &UNK& OS   & OS* &ALL &UNK& OS   & OS*   &ALL&UNK& OS   & OS*   &ALL&UNK\\\hline

OSVM &54.3&63.1&37.4&10.5&43.1&32.3&63.5&97.5&79.8&77.9&84.2&{\bf 89.0}&59.1&57.7&61.7&65.7\\	
MMD+OSVM &55.9& 64.7&39.1&12.2&62.8&58.9&69.5&82.1&80.0&79.8&81.3&81.0&68.0&68.8&66.3&58.4\\	
BP+OSVM &62.9&{\bf 75.3}&39.2&0.7&84.4&{\bf 92.4}&72.9&0.9&33.8&40.5&21.4&44.3&60.4&69.4&44.5&15.3\\\hline
Ours & {\bf 63.0}&59.1&{\bf 71.0} &{\bf 82.3}&{\bf 92.3}&91.2&{\bf 94.4}&{\bf 97.6}&{\bf92.1}&{\bf 94.9}&{\bf88.1}&78.0&{\bf 82.4}&{\bf 81.7}&{\bf 84.5}&{\bf 85.9}\\
\bottomrule[1.5pt]
\end{tabular}}
\caption{Accuracy (\%) of experiments on Digits Dataset.}
\vspace{-5mm}
\label{exp_digits}
\end{table}

\begin{table}[t!]
\centering
\scalebox{0.8}{
\begin{tabular}{c||cccc|cccc||cccc}
 \toprule[1.5pt]
& \multicolumn{4}{c}{MNIST}   &  \multicolumn{4}{|c|}{SVHN} &\multicolumn{4}{|c}{Average}\\
  \multicolumn{1}{c||}{Method}    &   OS   & OS*  &ALL &UNK& OS   & OS* &ALL &UNK& OS   & OS*   &ALL&UNK\\\hline
OSVM &85.4&82.9&90.3&97.8&77.7&79.7&75.7&68.1&81.5&81.2&83.0&82.9\\
Ours &{\bf 96.8}&{\bf 97.1}&{\bf 96.3}&95.5&72.3&82.9&60.7&19.2&84.5&{\bf 89.9}&78.5&57.3\\
Ours + OSVM &91.4&89.7&94.7&{\bf 99.7}&{\bf 87.3}&{\bf 86.1}&{\bf 89.0}&{\bf 93.1}&{\bf 89.3}&87.9&{\bf 91.8}&{\bf 96.3}\\
\bottomrule[1.5pt]
\end{tabular}
}
\caption{Accuracy (\%) on open set recognition experiments. Please note that domains of labeled and unlabeled samples are the same in this experiment.}
\label{op_setting}
\vspace{-3mm}
\end{table}
\vspace{-4mm}
\section{Conclusion}
\vspace{-3mm}
In this paper, we proposed a novel adversarial learning method for open set domain adaptation. Our proposed method enables the generation of features that can separate unknown target samples from known target samples, which is definitely different from existing distribution matching methods. 
Moreover, our approach does not require unknown source samples. 
Through extensive experiments, the effectiveness of our method has been verified. This method can be utilized to standard open set recognition as the result in Sec. \ref{std_opnst} demonstrates. Improving our method for the open set recognition will be our future work.
\section{Acknowledgements}
The work was partially supported by CREST, JST, and was partially funded by the ImPACT Program of the Council for Science, Technology, and Innovation
(Cabinet Office, Government of Japan). We would like to thank Kate Saenko for her great advice on our paper. 
\clearpage
\bibliographystyle{splncs}
\bibliography{egbib}
\appendix
\section{Detail of Experiments}
In this supplementary material, we describe the details of experiments which are not included in our main paper due to a limit of space. The experimental setting of baseline methods are provided in this material.
\subsection{Training Detail for Experiments on Office and VisDA Dataset}
\textbf{Experiments on Office}
With regard to the experiments on Office dataset, the batch-size was set as 32 and accuracy after 500 epoch was reported in all settings. 
Regarding to other baseline methods, MMD~\cite{long2015learning} and BP~\cite{ganin2014unsupervised}, the same architecture was utilized for training. With regard to these methods, since the performance dropped around 50 epochs, we reported the accuracy at 50 epochs. 
To obtain the results of MMD, we calculated the MMD between source and target features at the top layer of generator. We use 5 RBF kernels with the following standard deviation parameters:
\begin{equation}
  \sigma = [0.1,0.05,0.01,0.0001,0.00001].
\end{equation}
The training of BP was similar to our method, namely, the domain classifier was attached to the top layer of the feature generator networks. The original method utilized the learning rate decay to train a model. However, we did not utilize it because no improvement was observed. The training details of these baseline methods are almost the same for other experiments. The threshold value of open set SVM was fixed as 0.1, which was a default value of the SVM.

\textbf{Experiments on VisDA Dataset}
The accuracy after training 10 epoch was reported with regard to all methods. The other hyper-parameters are the same as the ones used in Office experiment. 

\textbf{Experiments on Digits Dataset}
We introduce the network architectures used in the experiments in Table \ref{tab:architecture}. Leaky ReLU was utilized for this experiments. The batch-size was set as 128 in this experiment. The accuracy after 200 epoch was reported in all settings. The same architecture was used for training other baselines OSVM, MMD and BP. 

\begin{table}[h!t]
\begin{minipage}[t]{.5\textwidth}
\footnotesize
\begin{center}
\begin{tabular}{|l|}
\hline
\multicolumn{1}{|c|}{SVHN $\rightarrow$ MNIST} \\
\hline
Input: $32 \times 32$ RGB image  \\\hline
\multicolumn{1}{|c|}{Generator} \\\hline
	Conv $5 \times 5 \times 64$, stride 1, pad 0\\    
    Conv $5 \times 5 \times 64$, stride 1, pad 0\\   
    Conv $3 \times 3 \times 128$, stride 2, pad 0\\  
	Conv $3 \times 3 \times 128$, stride 2, pad 0\\
	Fully connected $3200 \times 100$\\
	Fully connected $100 \times 100$\\\hline
    \multicolumn{1}{|c|}{Classifier}\\\hline
    Fully connected $100 \times 6$\\\hline
\end{tabular}
\end{center}
\end{minipage}
\begin{minipage}[t]{.5\textwidth}
\footnotesize
\begin{center}
\begin{tabular}{|l|}
\hline
\multicolumn{1}{|c|}{USPS$\leftrightarrow$MNIST} \\
\hline
Input: $28 \times 28$ Gray-scale image  \\\hline
\multicolumn{1}{|c|}{Generator} \\\hline
	Conv $5 \times 5 \times 20$, stride 1, pad 0\\    
    Max pooling $2 \times 2$, stride 2\\
    Conv $5 \times 5 \times 50$, stride 1, pad 0\\   
    Max pooling $2 \times 2$, stride 2\\
    Dropout 0.5\\
	Fully connected $800 \times 500$\\\hline
    \multicolumn{1}{|c|}{Classifier}\\\hline
    Fully connected $500 \times 6$\\\hline
\end{tabular}
\end{center}
\end{minipage}
\caption{Network architecture used for digits experiments. Leaky ReLU and Batch Normalization layer were used after every layer except for the classifier layer.}
\label{tab:architecture}
\end{table}

\section{Additional Experimental Results}
\begin{figure}[t]
\subfigure[Source Only (Source, Target Features)]{
   \includegraphics[width=0.4\hsize]{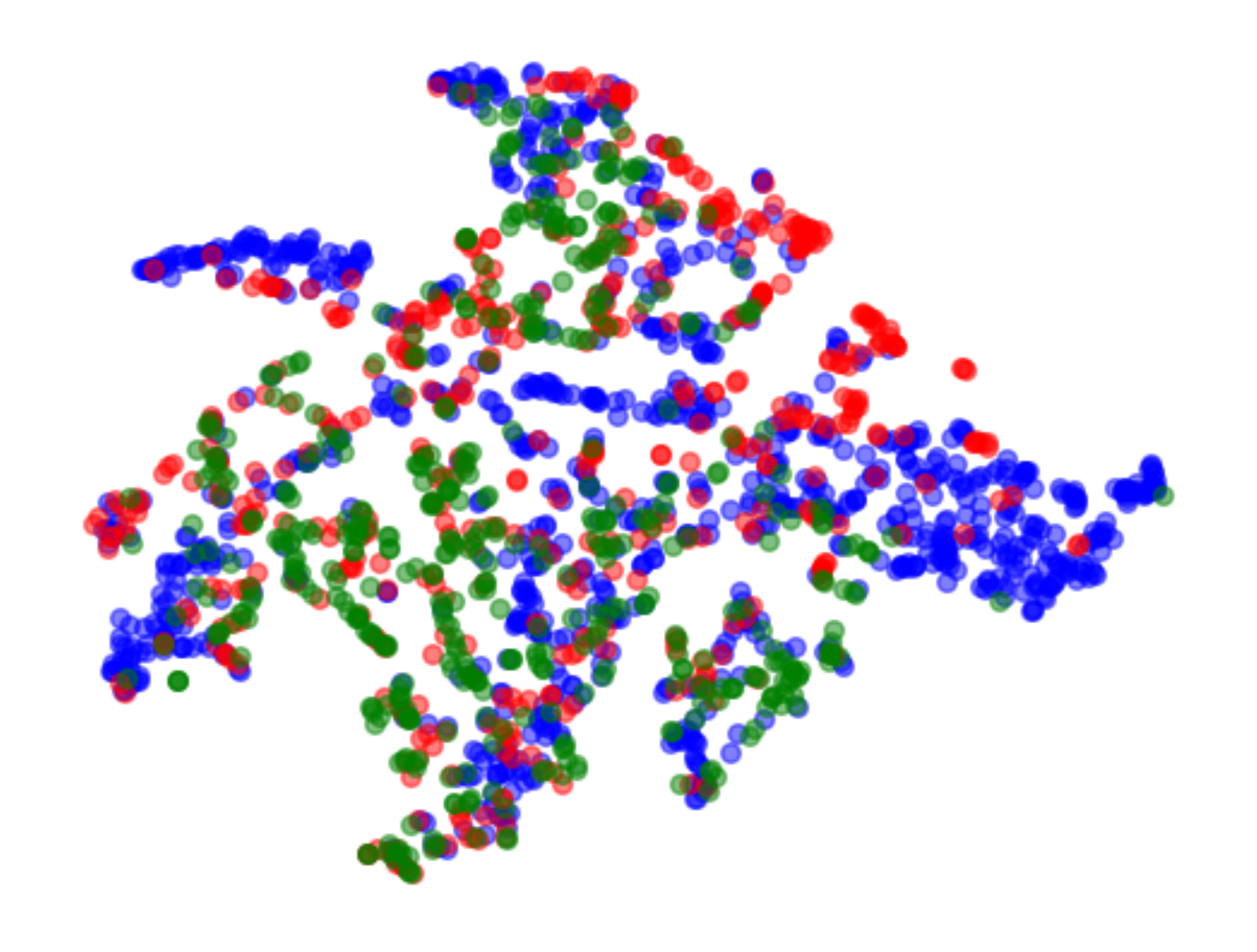}}
 \subfigure[Source Only (Target Features]{
  \includegraphics[width=0.5\hsize]{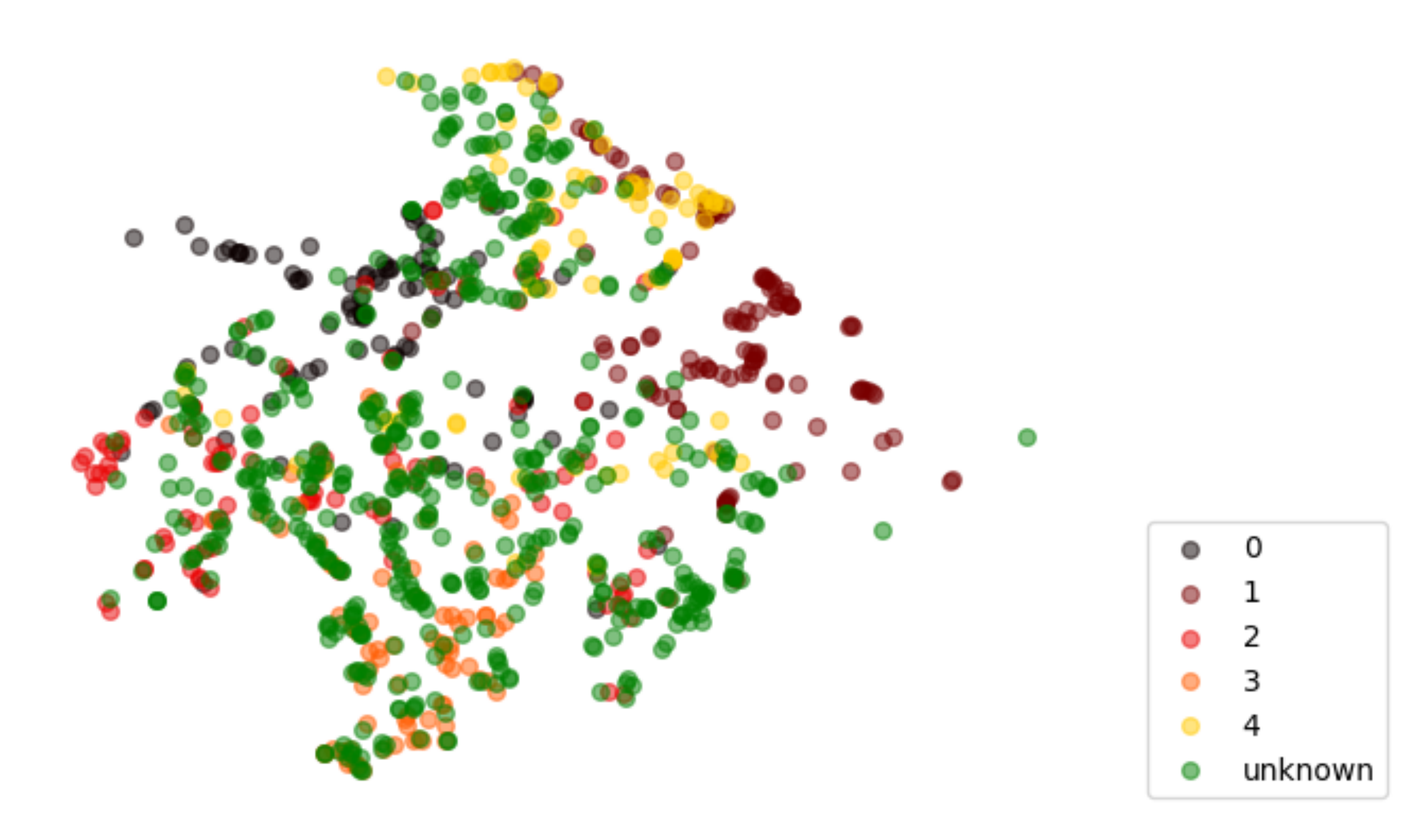}}
  \subfigure[MMD (Source, Target Features)]{
   \includegraphics[width=0.4\hsize]{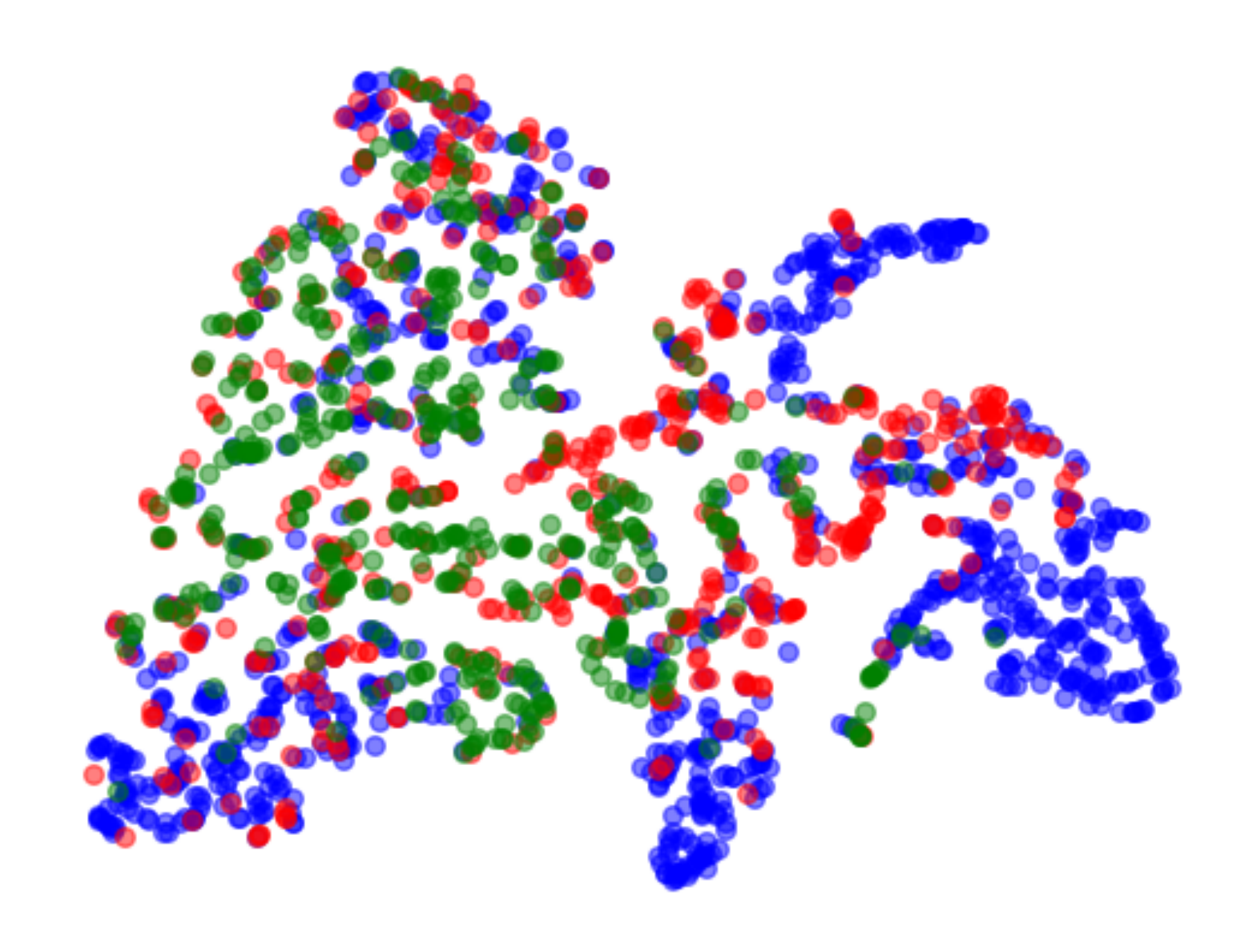}}
 \subfigure[MMD (Target Features)]{
  \includegraphics[width=0.5\hsize]{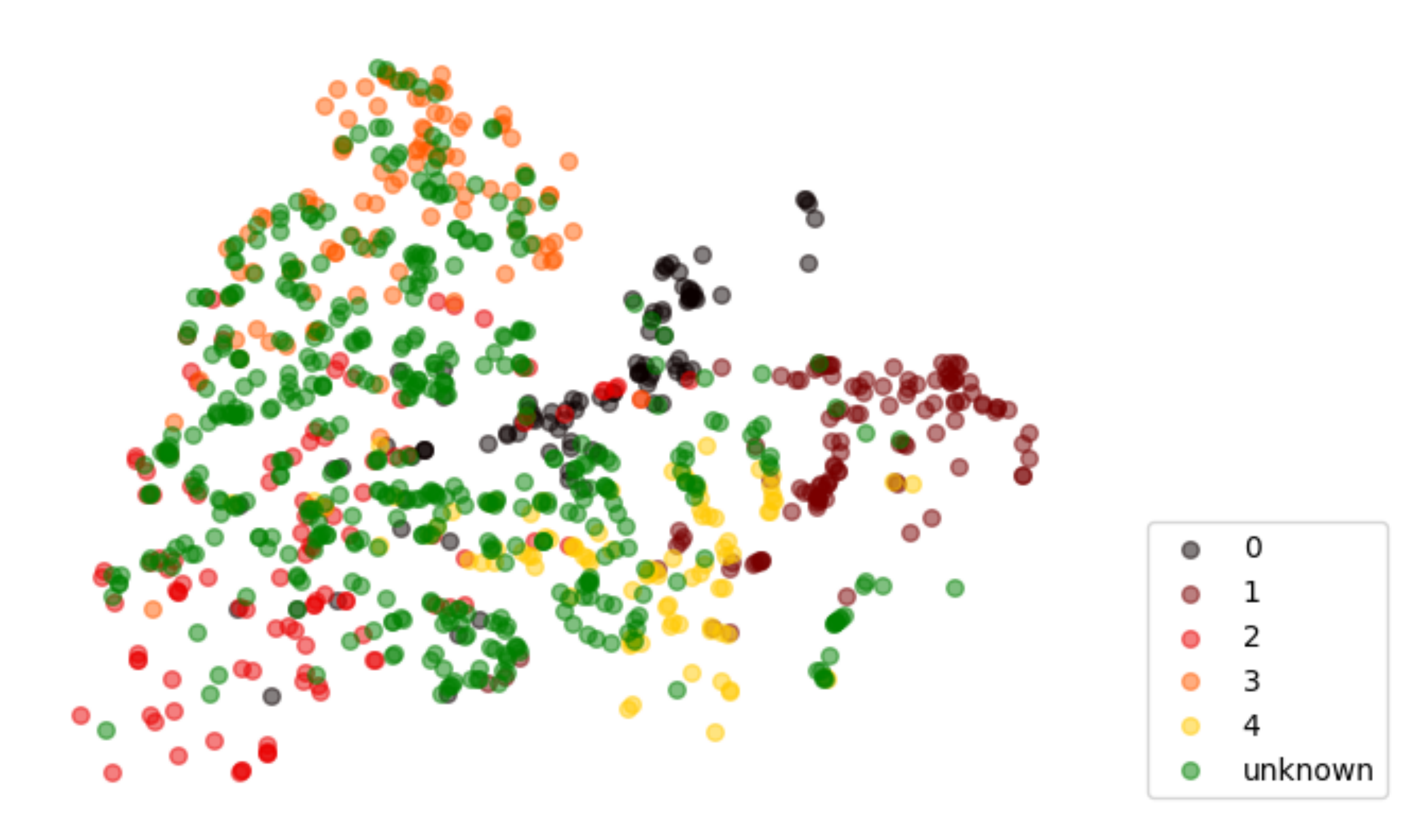}}
  \subfigure[BP (Source, Target Features)]{
   \includegraphics[width=0.4\hsize]{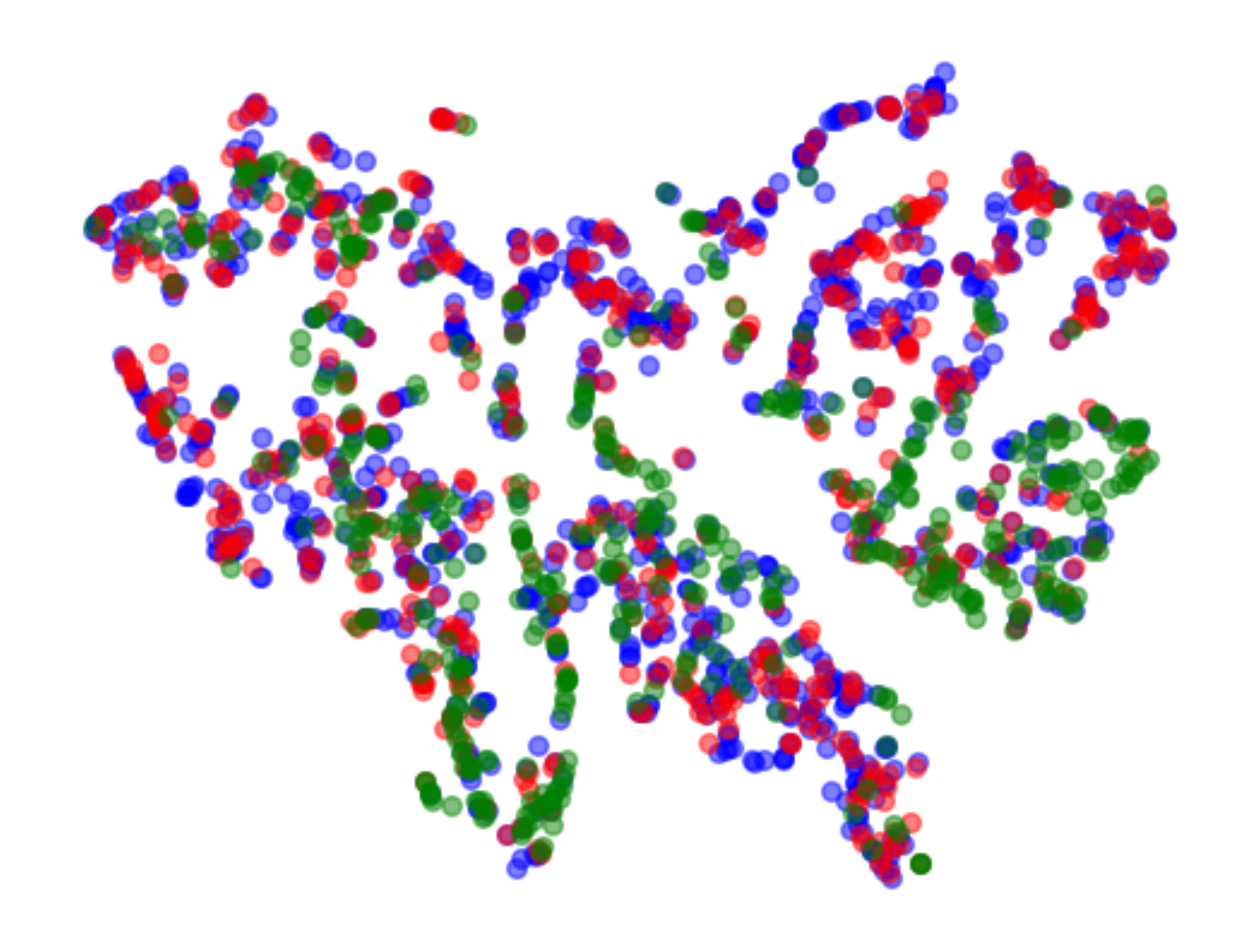}}
 \subfigure[BP (Target Features)]{
  \includegraphics[width=0.5\hsize]{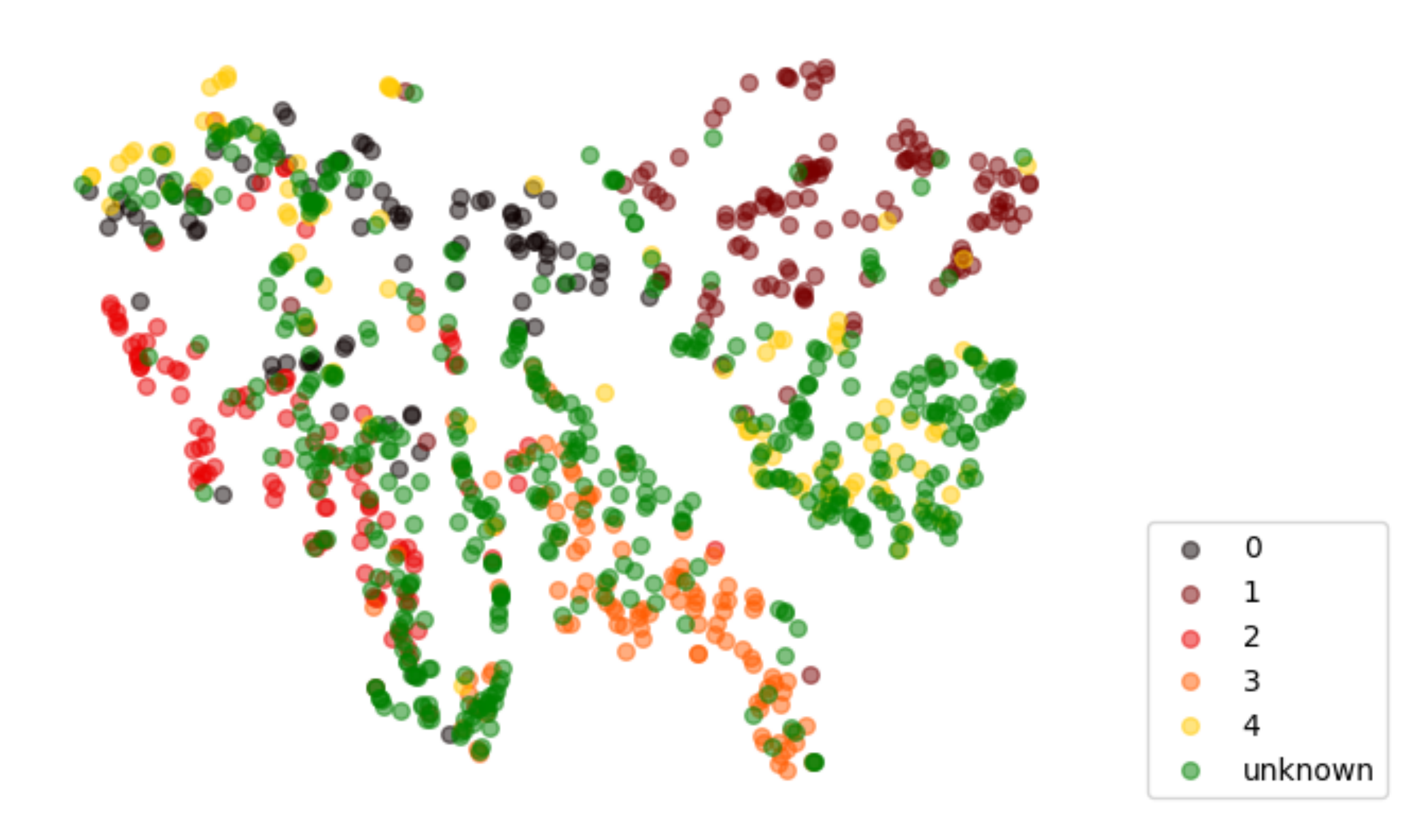}}
  \subfigure[Ours (Source, Target Features)]{
   \includegraphics[width=0.4\hsize]{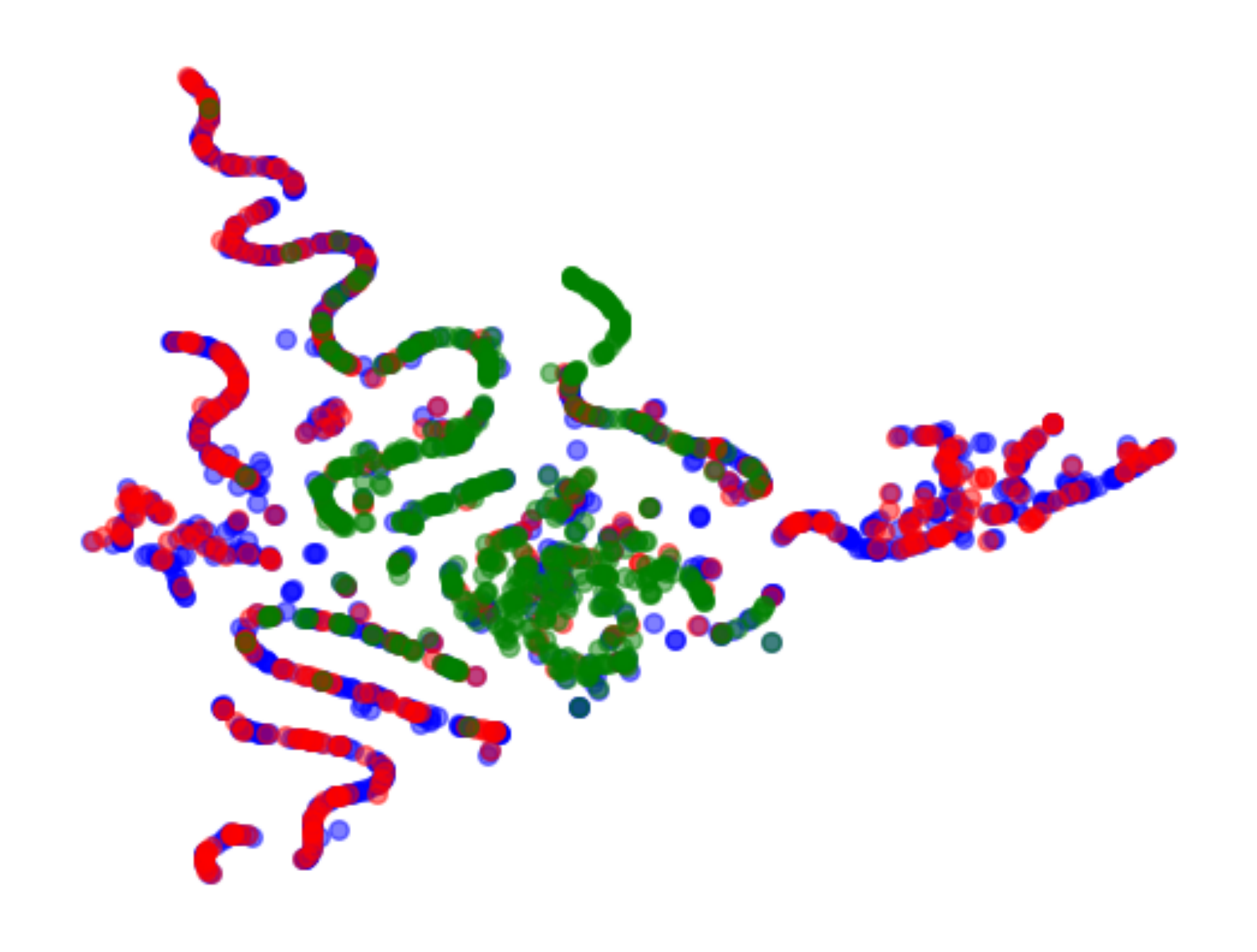}}
 \subfigure[Ours (Target Features)]{
  \includegraphics[width=0.5\hsize]{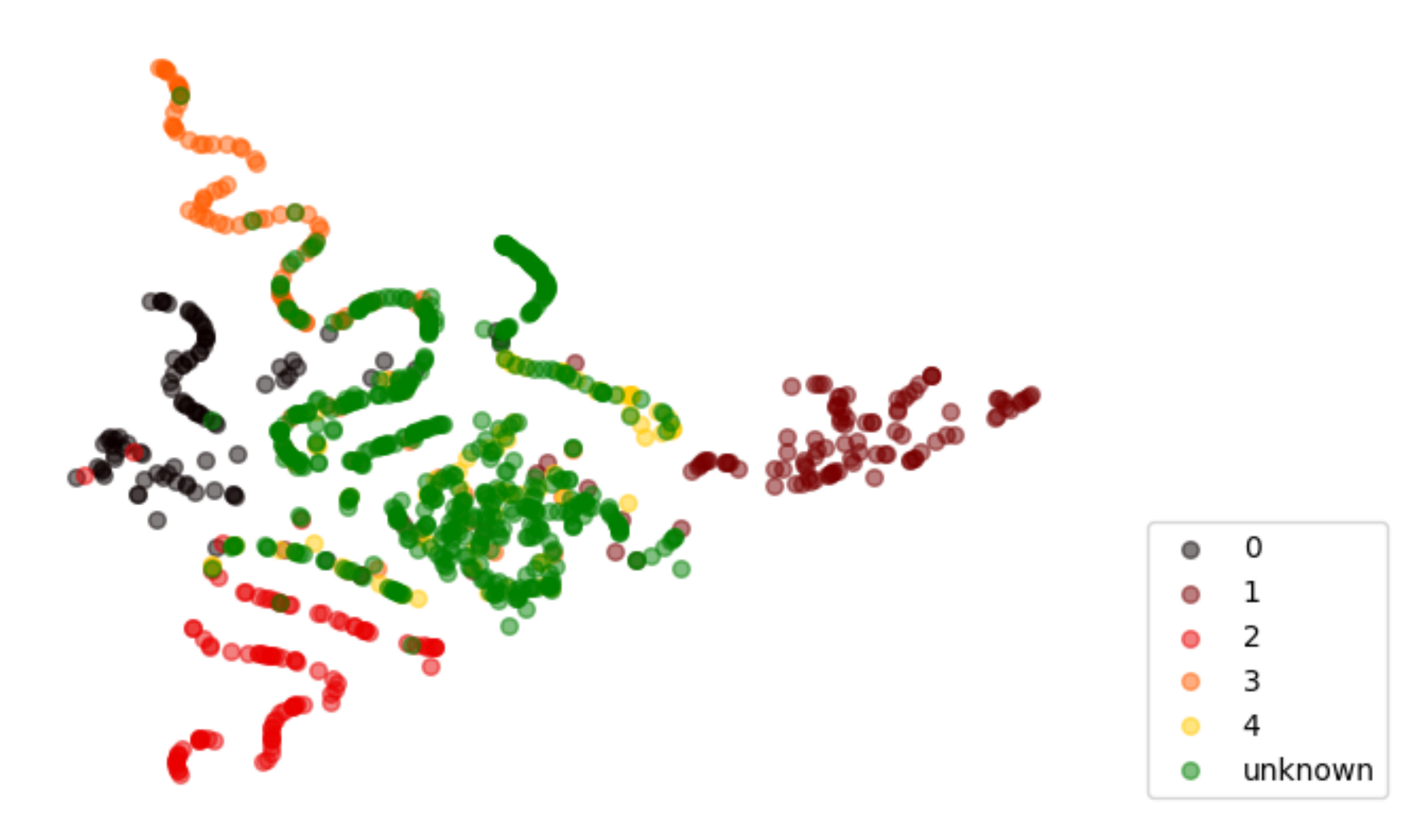}}
  
  \caption{Feature visualization of adaptation from SVHN to MNIST.}
  \label{vis:feature_svhn2mnist}
\end{figure}

Other results are shown in this section. 
Obtained features in adaptation from svhn to mnist  are visualized in Fig. \ref{vis:feature_svhn2mnist}. 
\end{document}